\documentclass{article}

    \PassOptionsToPackage{numbers, compress}{natbib}

\usepackage{preprint_2026}

\usepackage[utf8]{inputenc}
\usepackage[T1]{fontenc}
\usepackage{hyperref}
\usepackage{url}
\usepackage{booktabs}
\usepackage{amsfonts}
\usepackage{nicefrac}
\usepackage{microtype}
\usepackage{xcolor}
\usepackage{graphicx}
\usepackage{subcaption}
\usepackage{longtable}
\usepackage{makecell}
\usepackage[most]{tcolorbox}
\usepackage{amsmath}
\usepackage{amssymb}
\usepackage{mathtools}
\usepackage{amsthm}
\usepackage{multirow}
\usepackage{multicol}
\usepackage{tablefootnote}
\usepackage[capitalize,noabbrev]{cleveref}

\title{Retrieval-Augmented Linguistic Calibration}

\author{%
    Yi-Fan Yeh \\
    School of Computer Science \\
    University of Sydney\\
    Sydney, Australia \\
    \texttt{yyeh7345@uni.sydney.edu.au} \\
    \And
    Linwei Tao \\
    School of Computer Science \\
    University of Sydney\\
    Sydney, Australia \\
    \texttt{linwei.tao@sydney.edu.au} \\
    \And
    Minjing Dong \\
    City University of Hong Kong \\
    Hong Kong \\
    \texttt{minjdong@cityu.edu.hk} \\
    \And
    Tao Huang \\
    Shanghai Jiao Tong University \\
    Shanghai, China \\
    \texttt{t.huang@sjtu.edu.cn} \\
    \And
    Jialin Yu \\
    University of Oxford \\
    Department of Engineering Science \\
    Oxford, UK \\
    \texttt{jialin.yu@eng.ox.ac.uk} \\
    \And
    Philip Torr \\
    Department of Engineering Science\\
    University of Oxford \\
    Oxford, UK \\
    \texttt{philip.torr@eng.ox.ac.uk} \\
    \And
    Chang Xu \\
    School of Computer Science\\
    University of Sydney \\
    Sydney, Australia \\
    \texttt{c.xu@sydney.edu.au} \\
}

\begin{document}

\maketitle

\begin{abstract}
Linguistic cues such as ``I believe'' and ``probably'' offer an intuitive interface for communicating confidence, yet a generalisable, principled calibration framework for linguistic confidence expressions remains underexplored. In particular, co-occurring linguistic cues, contextual variation, and subjective audience interpretation pose unique challenges. We therefore model linguistic confidence as a distribution over plausible perceived probability values that a statement is correct, capturing interpretation variability that scalar representations discard. Within this distributional framework, we introduce faithfulness as a complementary evaluation dimension and present Faithfulness Divergence (FD), an information-theoretic metric quantifying the surprise induced in audience beliefs upon truth revelation. Building on these foundations, we present Retrieval-Augmented Linguistic Calibration (RALC), a lightweight post-hoc pipeline that propagates calibrated confidence signals back into natural language via retrieval-augmented rewriting. Across three QA benchmarks and five LLM families, RALC improves in-domain faithfulness and calibration up to 66\% and 58\%, respectively, outperforming black-box and grey-box calibration baselines.
\end{abstract}

\clearpage
\section{Introduction}

Reliable confidence estimation is fundamental to the trustworthy deployment of large language models (LLMs) in human decision-making pipelines \citep{Steyvers2025Whatlargelanguagemodelsknowandwhatpeoplethinktheyknow}. Without well-calibrated confidence signals, users risk over-relying on model outputs that hallucinate or fail silently \citep{kim2024examimpactofllms}, underscoring the need for confidence frameworks that are both scientifically rigorous and interpretable by human users.

Existing confidence estimation methods represent confidence as scalar probability values, including token-level probability \citep{lamb2025semantic}, semantic uncertainty \citep{gal2024semanticentropy}, and verbalised scores \citep{lin2022teachingmodelsexpressuncertainty, tian2023justaskcalibrationstrategies}. However, humans struggle to reason accurately with numerical probabilities \citep{zangmeloney2012probabilityperception}, motivating the use of linguistic markers such as ``may'' or ``likely'' as more natural confidence interfaces. Prior work demonstrates that such markers retain evaluative signal \citep{yona2024can, tao2025revisitinguncertaintyestimationcalibration}; however, treating them as scalars discards the inherent subjectivity of linguistic interpretation: different readers map the same expression to different perceived probability values \citep{tao2025largelanguagemodelsexpress}.

We address this gap by modelling linguistic confidence as a distribution over plausible perceived probability values that a statement is correct, where perception arises from readers' interpretations of the full linguistic content of the statement rather than from a discrete mapping of individual vocabulary items. Treating linguistic confidence as a surrogate for statement correctness induces a binary classification view, in which confidence corresponds to the predictive probability of the true class. Drawing a parallel to evidential deep learning, which models class probabilities with Dirichlet distributions to capture second-order predictive uncertainty \citep{sensoy2018evidentialdeeplearningquantify}, we formalise linguistic confidence for binary correctness using its natural binary special case: a Beta distribution over perceived confidence scores that a statement is correct, where the mean captures the central tendency of perceived confidence across readers and the concentration encodes the strength of agreement. 

The standard measure of confidence quality is calibration, assessed through population-level expected calibration error (ECE) \citep{guo2017calibrationmodernneuralnetworks, wang2025calibratingexpressionscertainty}, quantifying the alignment between confidence and accuracy in expectation. Instance-level metrics such as the Brier score \citep{verificationofforecastsexpressedintermsofprobability} and negative log likelihood offer pointwise assessment in the classical scalar setting, yet their distributional generalisations still fail to encode variance as the strength of agreement by readers. We therefore introduce \emph{faithfulness} as a complementary dimension of confidence evaluation and present Faithfulness Divergence (FD), a concentration-weighted Bayesian updating cost that quantifies the information-theoretic surprise to confidence beliefs upon truth revelation.

Calibration in linguistic space remains crucial yet largely unsolved. Classical post-hoc calibration methods, including temperature scaling \citep{guo2017calibrationmodernneuralnetworks}, Platt scaling \citep{PlattScaling1999}, histogram binning \citep{Zadrozny-histogram-binning-2001}, isotonic regression \citep{Zadrozny-isotonic-reg-2002}, Beta calibration \citep{pmlr-v54-kull17a}, and distribution-matching approaches \citep{song2019distributioncalibrationregression, marx2023calibrationdistributionmatchingtrainable}, operate entirely in numerical space and provide no mechanism for propagating calibrated signals back into language. Prompt-conditioned hedging strategies offer a linguistic alternative, yet function as black-box procedures with no principled control over the output \citep{yona2024can}. The closest related work performs discrete hedging word confidence profiling and remapping at the word level in a specialised domain \citep{wang2025calibratingexpressionscertainty}, overlooking the co-occurrence of multiple linguistic cues within a statement and their contextual interactions. Consequently, a generalisable, continuous, and lightweight post-hoc framework that provides principled guidance for hedging expressions remains underexplored. To address this gap, we introduce Retrieval-Augmented Linguistic Calibration (RALC), a post-hoc pipeline that operates directly in linguistic space to transform raw LLM responses into calibrated and faithful outputs. The pipeline applies Platt scaling \citep{PlattScaling1999} on confidence distribution means whilst preserving distributional concentration, and propagates calibrated distributions to language through retrieval-augmented LLM rewriting, employing the retrieval-augmented generation paradigm \citep{lewis2020retrieval}. Furthermore, RALC is compatible with diverse upstream confidence signals beyond linguistic confidence, including token probability and semantic uncertainty. 

Our contributions are as follows:
\begin{enumerate}
    \item We formalise linguistic confidence as a distribution over plausible perceived probability values that a statement is correct, capturing the interplay of linguistic cues and contexts beyond discrete expression mapping and scalar quantification.
    \item We introduce \emph{faithfulness} as a new dimension of confidence evaluation and present Faithfulness Divergence (FD), an instance-level metric that quantifies, as information-theoretic surprise, how faithfully a confidence distribution accounts for the ground-truth correctness outcome.
    \item We introduce a generalisable retrieval-augmented linguistic confidence calibration pipeline that effectively improves faithfulness and calibration in linguistic space and is compatible with diverse confidence estimation signals.
\end{enumerate}

We evaluate the framework across multiple LLM families and QA benchmarks, including MMLU \citep{hendrycks2021measuringmassivemultitasklanguage}, SQuAD~2.0 \citep{rajpurkar2018knowdontknowunanswerable}, and TruthfulQA \citep{lin2022truthfulqameasuringmodelsmimic}. Results demonstrate near-lossless information transfer through the calibration pipeline and substantial improvements in both calibration and faithfulness across models and benchmarks, outperforming the prompt-based calibration baselines. 

\section{Related work}

\paragraph{LLM confidence estimation}
Existing confidence estimation methods predominantly represent confidence as scalar probability values, including token-level probability aggregation \citep{lamb2025semantic, duan2024shiftingattentionrelevancepredictive} and consistency-based approaches that infer confidence from the semantic support landscape across repeated samples \citep{wang2023selfconsistencyimproveschainthought, gal2024semanticentropy}. Whilst recent work demonstrates that linguistic cues in model responses preserve evaluative signal as confidence surrogates \citep{yona2024can, tao2025largelanguagemodelsexpress}, their scalar quantification overlooks the inherently subjective nature of linguistic interpretation. \citet{wang2025calibratingexpressionscertainty} take a step towards distributional representations by mapping individual discrete hedging words to confidence distributions; however, their approach targets word-level remapping rather than statement-level confidence, where multiple linguistic cues co-occur and interact with context. \citet{huang2024calibratinglongformgenerationslarge} jointly model confidence and correctness as distributions over ambiguous long-form generation contexts, which is orthogonal to our objective of binary classification with predictive probability distributions. 

\paragraph{Confidence evaluation}
Expected Calibration Error (ECE) is the dominant metric for evaluating confidence, measuring alignment between scalar confidence scores and accuracy in both classical \citep{guo2017calibrationmodernneuralnetworks} and language model settings \citep{zhu-etal-2023-calibration}. Extensions based on entropy \citep{sumler2025entropicmetricmeasuringcalibration}, variance \citep{thompson2026extendingconfidencecalibrationgeneralised}, and distributional generalisation \citep{wang2025calibratingexpressionscertainty} remain ECE-based and rely on local aggregation, discarding full distributional information at the instance level. Instance-level scoring such as the Brier score \citep{verificationofforecastsexpressedintermsofprobability} and negative log likelihood similarly do not capture variance-scaled misalignment. 

\paragraph{Confidence calibration}
Classical post-hoc calibration methods, including Platt scaling \citep{PlattScaling1999}, histogram binning \citep{Zadrozny-histogram-binning-2001}, isotonic regression \citep{Zadrozny-isotonic-reg-2002}, and Beta calibration \citep{pmlr-v54-kull17a}, adjust scalar outputs towards empirical accuracy but are confined to numerical space. Distributional calibration methods frame the problem as distribution matching, aligning predicted confidence distributions with empirical label distributions through various mapping strategies \citep{song2019distributioncalibrationregression, marx2023calibrationdistributionmatchingtrainable}, though they target global rather than instance-level calibration. In the linguistic space, prompt-based strategies have been explored to steer LLM hedging \citep{yona2024can}, but these lack principled control. Internal model steering offers finer-grained calibration of verbal uncertainty \citep{ji2025calibratingverbaluncertaintylinear}, yet requires access to model internals, limiting applicability to open-source settings. The most closely related approach remaps discrete hedging words at the vocabulary level, without accounting for contextual interactions or producing calibrated full responses \citep{wang2025calibratingexpressionscertainty}.

\section{Confidence estimation and evaluation}
\label{section:confidence-estimation}

\subsection{Linguistic confidence estimation}
\label{section:linguistic-confidence-estimation}
For each input--response pair $(X, R)$, let $y \in \{0,1\}$ denote the correctness label of $R$. We define a distributional confidence estimator
\[
g : \mathcal{R} \rightarrow \mathcal{P}([0,1]),
\] 
where $\mathcal{R}$ denotes the space of model responses and $\mathcal{P}([0,1])$ denotes the space of probability distributions over $[0,1]$. The estimator $g$ models the plausible probability values that $R$ is correct as a distribution over confidence scores in $[0,1]$ as perceived by readers (human or model-based evaluators). We abstract $g$ as a model-based or human-based evaluator and parameterise the estimated distribution $S$ as a Beta distribution, $S = \mathrm{Beta}(\alpha, \beta)$. This choice draws a parallel to evidential deep learning \citep{sensoy2018evidentialdeeplearningquantify}, which places a Dirichlet prior over class probabilities to represent second-order uncertainty. Our setting resembles binary classification: each reader produces an interpreted confidence score viewable as a draw of the true-class probability; the Beta distribution is therefore the principled choice, as the binary special case of the Dirichlet and the natural conjugate prior for the Bernoulli likelihood. The mean $\alpha/(\alpha+\beta)$ captures the central tendency of perceived confidence across readers, whilst the concentration $(\alpha+\beta)$ encodes agreement strength: a high mean with low concentration signals inconsistent reader interpretations, whereas the same mean with higher concentration signals consistent agreement.

\subsection{Confidence evaluation}

\paragraph{Calibration as a dimension of confidence evaluation}
At the population level, calibration requires confidence to match empirical accuracy in expectation. Letting $p \sim S$ denote the scalar confidence value drawn from the estimated distribution, the classical measure is the expected calibration error,
\[
\text{ECE} = \mathbb{E}\!\left[\left|\mathbb{E}[Y \mid p] - p\right|\right],
\]
with realisations including scalar bin-based \citep{guo2017calibrationmodernneuralnetworks} and distribution-generalised variants \citep{wang2025calibratingexpressionscertainty}.

\paragraph{Faithfulness as a dimension of confidence evaluation}
Calibration is necessary but not sufficient in the distributional setting: two predictors may achieve similar average calibration yet convey markedly different instance-level confidence profiles. We therefore introduce \emph{faithfulness}, a human-aligned, instance-level dimension of confidence evaluation.

A confidence distribution is faithful when observing the ground truth induces little surprise relative to prior beliefs. Such surprise is driven by both central-tendency misalignment and the strength of agreement with which that misalignment is held. Grounding this intuition in information theory, we draw on Bayesian surprise \citep{itti2009bayesian}, measured by the KL divergence between posterior and prior distributions, and weight it by the concentration $(\alpha_i + \beta_i)$ as the effective sample size of the prior \citep{morita2007determining} to represent the total surprise of the update.

Formally, for instance $i$, we update the prior $S_i$ with a single Bernoulli observation of the correctness label $y_i$ to obtain the posterior $S_i^*$:
\[
S_i = \mathrm{Beta}(\alpha_i,\, \beta_i), \qquad S_i^* = \mathrm{Beta}(\alpha_i + y_i,\; \beta_i + 1 - y_i).
\]
The \emph{Faithfulness Divergence} (FD) for instance $i$ is then
\[
\mathrm{FD}_i := (\alpha_i + \beta_i)\cdot\mathrm{KL}\!\left(S_i^* \,\|\, S_i\right).
\]
The KL term quantifies the information-theoretic change of belief required upon observing the outcome; weighting by $(\alpha_i + \beta_i)$ ensures the significance of the change is properly accounted for. FD is non-negative and intended as a \emph{relative} instance-level metric: lower values indicate more faithful confidence communication, whilst higher values signal greater mismatch between expressed confidence and realised correctness. We provide a further discussion of the modelling in Appendix~\ref{appendix:faithfulness-discussion}.

\section{Retrieval-Augmented Linguistic Calibration (RALC)}
\label{section:ralc}

\begin{figure}
    \centering
    \includegraphics[width=\linewidth]{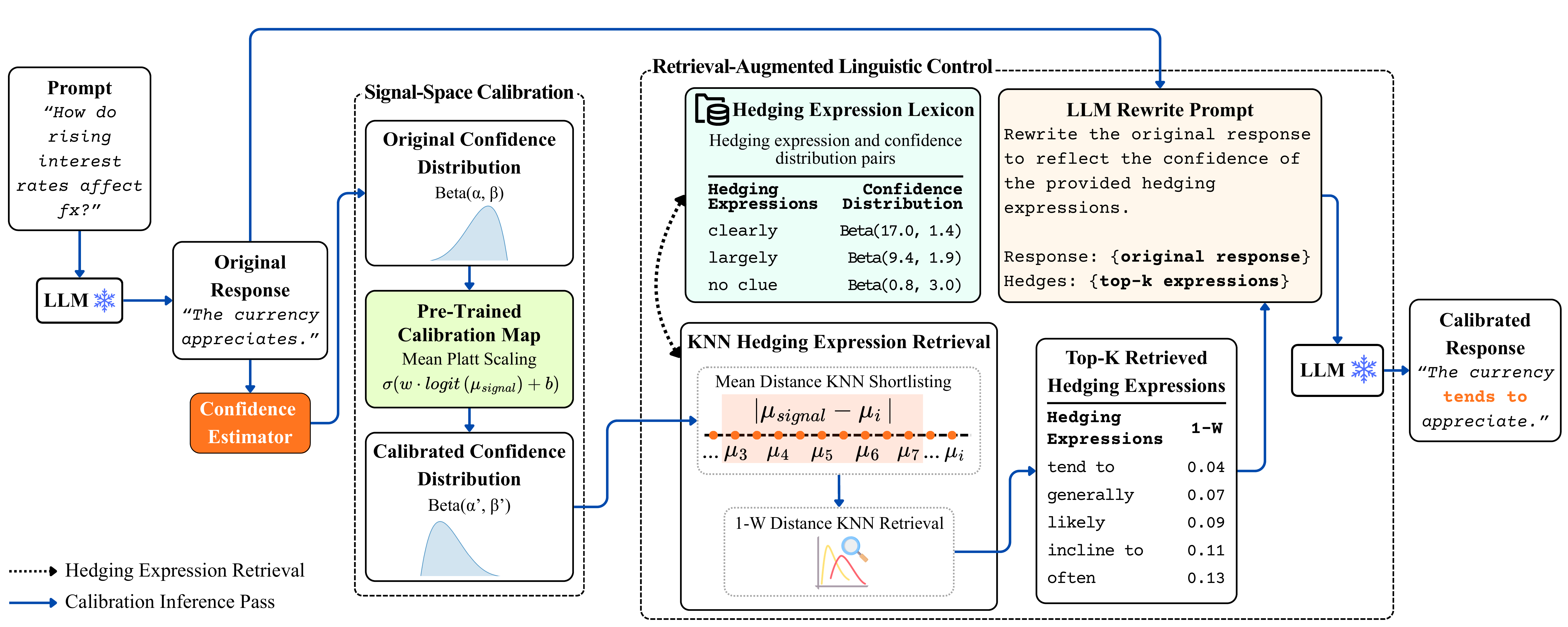}
    \caption{Retrieval-Augmented Linguistic Calibration pipeline overview. In each calibration inference pass (\textbf{\textcolor{blue!90!black}{blue arrow $\rightarrow$}}), we estimate a confidence distribution for the original response using linguistic confidence, token probability, or semantic uncertainty (Sections~\ref{section:linguistic-confidence-estimation}, \ref{section:alternative-confidence-signals}). In the signal space, we apply a pre-trained Platt scaling calibration map on the mean to correct miscalibration in the numerical space (Section~\ref{section:post-hoc-signal-space-calibration}). The calibrated distribution is then used as a retrieval signal to find the nearest hedging expressions from a pre-built hedge-confidence-pair lexicon. The KNN retrieval process uses absolute distance in the means for shortlisting and 1-Wasserstein distance for final retrieval to ensure alignment in both central tendency and spread. The retrieved hedging expressions form a rewrite prompt along with the original response, which is passed to an LLM editor to produce a linguistically calibrated response (Section~\ref{section:retrieval-augmented-linguistic-control}).}
    \label{fig:linguistic-calibration-pipeline-infograph}
\end{figure}

We introduce a novel post-hoc calibration pipeline that operates directly in the linguistic space to transform original LLM responses into calibrated and faithful outputs, Retrieval-Augmented Linguistic Calibration (RALC). RALC transforms a raw response into one whose perceived confidence more faithfully reflects the underlying ground-truth label. Figure~\ref{fig:linguistic-calibration-pipeline-infograph} illustrates the pipeline mechanics. It estimates a confidence distribution for the original responses, applies post-hoc signal-space calibration to correct miscalibration in the numerical space, and then uses the calibrated distribution as a retrieval signal to search for appropriate hedging expressions that guide the rewriting of the original response into a linguistically calibrated and faithful one.

\subsection{Post-hoc signal-space calibration}
\label{section:post-hoc-signal-space-calibration}
\paragraph{Definition}
Under the distributional setting, the confidence estimator $g$ outputs a
distribution over confidence scores. Post-hoc calibration learns a mapping
$t$ in distribution space so that the calibrated estimator $t \circ g$
reduces expected calibration error and Faithfulness Divergence.

\paragraph{Platt scaling on distribution means}
We parameterise confidence with a Beta distribution
$S = \mathrm{Beta}(\alpha, \beta)$, with mean $\mu = \alpha/(\alpha+\beta)$
and concentration $\kappa = \alpha + \beta$. We apply Platt scaling
\citep{PlattScaling1999} on the distribution mean, mapping $\mu$ to a
calibrated mean $\mu'$ by fitting a logistic regression of distribution
means against binary correctness labels:
\begin{equation}
    \mu' = \sigma\!\left(w \cdot \operatorname{logit}(\mu) + b\right),
\end{equation}
where $\sigma(\cdot)$ denotes the sigmoid function and $(w, b) \in \mathbb{R}^2$
are learned scalar parameters. The concentration $\kappa = \alpha + \beta$ is
preserved from the original estimated distribution due to the non-existence of a natural target for concentration calibration. 

We reconstruct the calibrated Beta distribution from the calibrated mean
$\mu'$ and the preserved concentration $\kappa$, setting
$\alpha' = \mu' \cdot \kappa$ and $\beta' = (1 - \mu') \cdot \kappa$,
yielding the calibrated distribution $t(S) = \mathrm{Beta}(\alpha', \beta')$. We also investigate classical alternatives, including temperature scaling \citep{guo2017calibrationmodernneuralnetworks}, isotonic regression \citep{Zadrozny-isotonic-reg-2002}, and histogram binning \citep{Zadrozny-histogram-binning-2001}, and find that Platt scaling is the consistent outperformer in improving calibration and faithfulness. We present an ablation study in Appendix~\ref{appendix:calibration-map-comparison}.

\paragraph{Interpretation}
The signal-space Platt scaling corrects systematic mean misalignment, whilst preserving the concentration that encodes the strength of agreement across readers. The mean is directly supervised by the correctness labels and scaling induces a more faithful and calibrated belief of the correctness outcome. By contrast, the concentration is a linguistic feature that emerges from the interplay of linguistic cues and reader interpretation, and does not have a theoretical target for calibration. As a result, we deliberately confine our calibration design to one mean scaling where a natural target exists. 

\subsection{Retrieval-augmented linguistic control}
\label{section:retrieval-augmented-linguistic-control}
Post-hoc signal-space calibration updates the originally estimated confidence distributions but leaves the response language unchanged. To close this gap, we introduce retrieval-augmented linguistic control, which rewrites the original response $R$ into a revised response $R'$ whose perceived confidence aligns with the calibrated signal $S' = t(S)$. Formally, a linguistic calibrator $l$ produces $R' = l(R, S')$ such that re-estimating confidence from $R'$ recovers $S'$. The full pipeline $l \circ t \circ g$ thus forms a closed loop from raw response to calibrated confidence to linguistically calibrated response. 

\paragraph{Linguistic confidence lexicon}
The retrieval step relies on a lexicon that maps hedging expressions to confidence distributions. Hedging expressions are sourced from state-of-the-art LLMs, including Claude-Sonnet-4.6 \citep{anthropic2026sonnet46}, GPT-5.4 \citep{openai2026gpt54}, and Gemini-3-Flash \citep{googledeepmind2026gemini3flash}. For each hedging expression $w_k$, GPT-OSS-20B \citep{agarwal2025gpt} rewrites a collection of non-verifiable statements to incorporate that expression. The LLM linguistic evaluator ensemble then independently evaluates the perceived confidence of each rewritten statement in 3 model passes per evaluator, producing a set of confidence scores, as outlined in Section~\ref{section:setup-preliminaries}. Fitting a Beta distribution to all confidence scores across passes yields the pair $\bigl(w_k,\, \mathrm{Beta}(\alpha_k, \beta_k)\bigr)$. Repeating this procedure across all hedging expressions produces the lexicon $\{(w_k,\, \mathrm{Beta}(\alpha_k, \beta_k))\}_{k=1}^{K}$ used at inference time. Additional details on the lexicon construction are provided in Appendix~\ref{appendix:lexicon-construction}.

\paragraph{Retrieval-based rewriting}
Given a calibrated signal $S'$, we retrieve the $k$ nearest hedging expressions from the lexicon via a two-stage process. First, we shortlist candidates by mean distance $|\mu_k - \mu_{S'}|$, retaining expressions whose distributional mean falls within a neighbourhood of $\mu_{S'}$. Second, we rank the shortlisted candidates by the $1$-Wasserstein distance via Monte Carlo estimation,
\[
d(w_k,\, S') = W_1\!\left(\mathrm{Beta}(\alpha_k, \beta_k),\, S'\right),
\]
and select the top-$k$ nearest expressions. Mean-distance shortlisting efficiently narrows the candidate set, whilst the subsequent $W_1$ ranking captures full distributional shape, matching both the central tendency and spread of $S'$, at lower computational cost than applying $W_1$ over the entire lexicon. The retrieved expressions are then passed alongside $R$ to an LLM editor, which rewrites $R$ into $R'$ to match the target confidence profile, enforcing $g(R') \approx S'$ in practice. 

\subsection{Alternative confidence signals}
\label{section:alternative-confidence-signals}
The RALC pipeline is compatible with confidence signals beyond linguistic confidence, including token probability \citep{lamb2025semantic} and semantic uncertainty \citep{gal2024semanticentropy}, serving as the post-hoc signal-space calibration object and retrieval signal to guide the calibration process. We formulate both distributional confidence signals under self-consistency sampling \citep{wang2023selfconsistencyimproveschainthought} and cluster the generated responses semantically. 

\paragraph{Length-normalised token probability}
Let $\mathcal{I}_{\max}$ denote the index set of the largest cluster. For each $R_j \in \mathcal{I}_{\max}$, where $r_i$ is the $i$-th token and $r_{<i}$ the preceding context, we compute its length-normalised token probability \citep{lamb2025semantic}:
\[
s_j^{\mathrm{tok}} = \exp\!\left(\frac{1}{|R_j|}\sum_{i=1}^{|R_j|}\log p_\theta(r_i \mid r_{<i}, X)\right) \in [0,1].
\]
Fitting a Beta distribution to $\{s_j^{\mathrm{tok}}\}$ via method of moments (Appendix~\ref{appendix:beta-method-of-moments}) yields $S^{\mathrm{tok}}$ that represents confidence as a distribution of token-level probabilities for a particular semantic meaning.

\paragraph{Semantic uncertainty}
Let $\mathcal{I}_{\max}$ denote the index set of the largest cluster and $N$ be the total number of sampled responses. For each self-consistency sample for a given input, the Beta parameters are set directly from cluster counts,
\[
\alpha^{\mathrm{sem}} = \lvert\mathcal{I}_{\max}\rvert, \qquad \beta^{\mathrm{sem}} = N - \lvert\mathcal{I}_{\max}\rvert,
\]
Both parameters are clipped to a minimum of $10^{-6}$ to handle degenerate cases (e.g.\ all responses falling into a single cluster, which would set $\beta^{\mathrm{sem}} = 0$), yielding $S^{\mathrm{sem}} = \mathrm{Beta}(\max(\alpha^{\mathrm{sem}}, 10^{-6}),\, \max(\beta^{\mathrm{sem}}, 10^{-6}))$ that represents confidence as a distribution of semantic support for a particular semantic meaning across samples.

\begin{table}[t]
\centering
\small
\caption{Instance-level metric comparison across controlled subsets varying in mean confidence, concentration, and accuracy. Only FD correctly ranks surprise levels consistent with each subset's distributional profile; KL divergence, $\mathbb{E}[\text{Brier}]$, and $\mathbb{E}[\text{NLL}]$ each fail to recover the expected ordering.}
\label{tab:fd_miscalibration}
\begin{tabular}{lccccccc}
\toprule
\textbf{Subset} & \textbf{Acc.} & \textbf{Avg Conf.} & \textbf{Conc.} & \textbf{FD}$\downarrow$ & \textbf{KL} & \textbf{$\mathbb{E}[\text{Brier}]$} & \textbf{$\mathbb{E}[\text{NLL}]$} \\
\midrule
(1) high conf., high conc., wrong & 0.0 & 0.812 & 25.8 & 2.932 & 0.168 & 0.681 & 2.052 \\
(2) low conf.,  high conc., right & 1.0 & 0.448 & 6.0 & 0.550 & 0.114 & 0.348 & 0.953 \\
(3) high conf., low conc., wrong & 0.0 & 0.656 & 1.0 & 0.486 & 0.590 & 0.557 & 3.832 \\
(4) low conf.,  low conc., right & 1.0 & 0.433 & 1.0 & 0.392 & 0.378 & 0.442 & 1.719 \\
\bottomrule
\end{tabular}
\end{table}

\section{Experiments}
\label{section:experiments}

\subsection{Setup preliminaries}
\label{section:setup-preliminaries}

We evaluate across five open-source language models from different families: GPT-OSS-20B \citep{agarwal2025gpt}, Llama-3.1-8B-Instruct \citep{meta2024llama31}, Qwen3-8B \citep{qwen3}, Mistral-7B-Instruct-v0.3 \citep{mistralai2023mistral7b}, and Gemma-4-31B-IT \citep{googledeepmind2025gemma4}, on three benchmarks: MMLU \citep{hendrycks2021measuringmassivemultitasklanguage}, SQuAD~2.0 \citep{rajpurkar2018knowdontknowunanswerable}, and TruthfulQA \citep{lin2022truthfulqameasuringmodelsmimic}, covering reasoning-heavy multiple-choice, reading comprehension, and closed-book short-answer question-answering formats. We elicit responses using the Direct QA and Hedged QA prompt templates based on \citet{yona2024can}'s work. Direct QA aims to produce natural, succinct free-form responses, whilst Hedged QA additionally instructs the model to express uncertainty through hedging language and serves as a black-box baseline for RALC. The full templates are in Appendix~\ref{appendix:qa-prompts}. We employ GPT-OSS-20B as a model-based grader for correctness and the rewriting model in our RALC pipeline. To estimate linguistic confidence as perceived by readers, we construct an evaluator ensemble of three LLMs (Qwen3-8B, Llama-3.1-8B-Instruct, and Mistral-7B-Instruct-v0.3) as a human audience surrogate, capturing potential co-occurrences of linguistic cues and their contextual interactions; each model independently evaluates the confidence expressed in a response three times. A Beta distribution is fitted to these scores via method of moments (Appendix~\ref{appendix:beta-method-of-moments}). We further validate our LLM ensemble against the human-annotated linguistic benchmark of \citet{tao2025largelanguagemodelsexpress} (Appendix~\ref{appendix:llm-ensemble-human-benchmark}).

\begin{figure}
    \centering
    \includegraphics[width=0.9\linewidth]{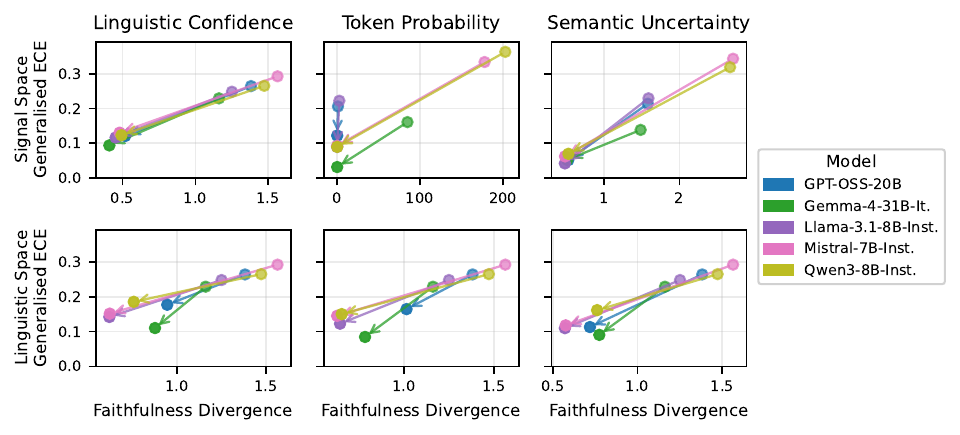}
    \caption{Pre-calibration$\to$post-calibration changes in generalised ECE and Faithfulness Divergence across signal space (\textbf{top}) and linguistic space (\textbf{bottom}), averaged across MMLU, SQuAD~2.0, and TruthfulQA. Our RALC consistently improves (reduces) both metrics across all confidence signals in both spaces.}
    \label{fig:in-domain-calibration-direct-qa}
\end{figure}

\subsection{Measuring faithfulness under distributional confidence}

To validate the quantification of the \emph{surprise after truth revelation}, we construct four controlled subsets from the linguistic confidence distributions of Llama-3.1-8B-Instruct on SQuAD~2.0, spanning the corners of the distributional confidence space by varying mean confidence, concentration, and accuracy. We compare FD against KL divergence \citep{lk-divergence-1951}, $\mathbb{E}[\text{Brier}]$ \citep{verificationofforecastsexpressedintermsofprobability}, and $\mathbb{E}[\text{NLL}]$. Whilst the Brier score and NLL are commonly used in scalar settings, we provide their distributional generalisations in Appendix~\ref{appendix:distributional-brier-nll}.

As shown in Table~\ref{tab:fd_miscalibration}, FD is the only metric that recovers the expected surprise ordering: concentrated, misaligned distributions receive the highest penalties (surprise), whilst the same misalignment expressed with low concentration receives lower penalties. KL divergence inverts this ranking by penalising diffuse distributions more heavily regardless of mean misalignment due to a lack of weighting by the effective sample size. $\mathbb{E}[\text{Brier}]$ and $\mathbb{E}[\text{NLL}]$ also fail to recover the expected surprise ordering, as neither encodes variance as an amplifier or mediator of surprise. These results confirm that FD uniquely captures both mean misalignment and dispersion, making it the appropriate instance-level faithfulness metric in the distributional confidence setting under our information-theoretic modelling. In addition to the empirical validation, we provide further theoretical ablation studies on FD in Appendix~\ref{appendix:additional-fd-ablations}.

\subsection{Retrieval-Augmented Linguistic Calibration (RALC)}

\subsubsection{In-domain calibration}

\paragraph{Signal-space and linguistic-space calibration}

We evaluate RALC's ability to improve calibration and faithfulness across both signal and linguistic spaces. For each question, we generate 20 responses under self-consistency sampling, cluster them semantically, and identify the largest cluster. The first response in the majority cluster is selected as the original response for linguistic calibration. Linguistic confidence is estimated by our LLM ensemble on this response; token probability and semantic uncertainty distributions are constructed from the cluster according to Section~\ref{section:alternative-confidence-signals}. We provide the LLM-based clustering prompt in Appendix~\ref{appendix:semantic-clustering-prompt}.

For each confidence signal, we train a Platt scaling calibration map on the first 30\% of each dataset, regressing per-response distribution means against binary correctness labels $y \in \{0, 1\}$. We then run RALC inference on the remaining 70\% held-out set, applying the pre-trained Platt scaling map to the estimated confidence distributions to obtain calibrated distributions. The calibrated distributions are passed to the retrieval-augmented linguistic control module, which selects the top $k{=}5$ from the top 30 mean-based shortlisted hedging expressions to rewrite the original response into a linguistically calibrated one. We provide the LLM rewriting prompt in Appendix~\ref{appendix:retrieval-augmented-linguistic-calibration} and the choice of $k$ ablation study in Appendix~\ref{appendix:top-k-retrieval-ablation}.

We evaluate pre- and post-calibration in the signal space by comparing the Faithfulness Divergence and generalised ECE \citep{wang2025calibratingexpressionscertainty} of the confidence distributions before and after the calibration map. In the linguistic space, we estimate the linguistic confidence for both the original and linguistically calibrated responses using the LLM ensemble and evaluate using the same metrics. As shown in Figure~\ref{fig:in-domain-calibration-direct-qa}, RALC consistently improves both metrics across all three confidence signals in both spaces. Additionally, across all signals, semantic uncertainty yields the strongest improvements in calibration and faithfulness as shown in Table~\ref{tab:cross-domain-calibration}. Additional results are detailed in Appendix~\ref{appendix:additional-results}.

\paragraph{Benchmark against calibration baselines}

\begin{figure}
    \centering
    \includegraphics[width=\linewidth]{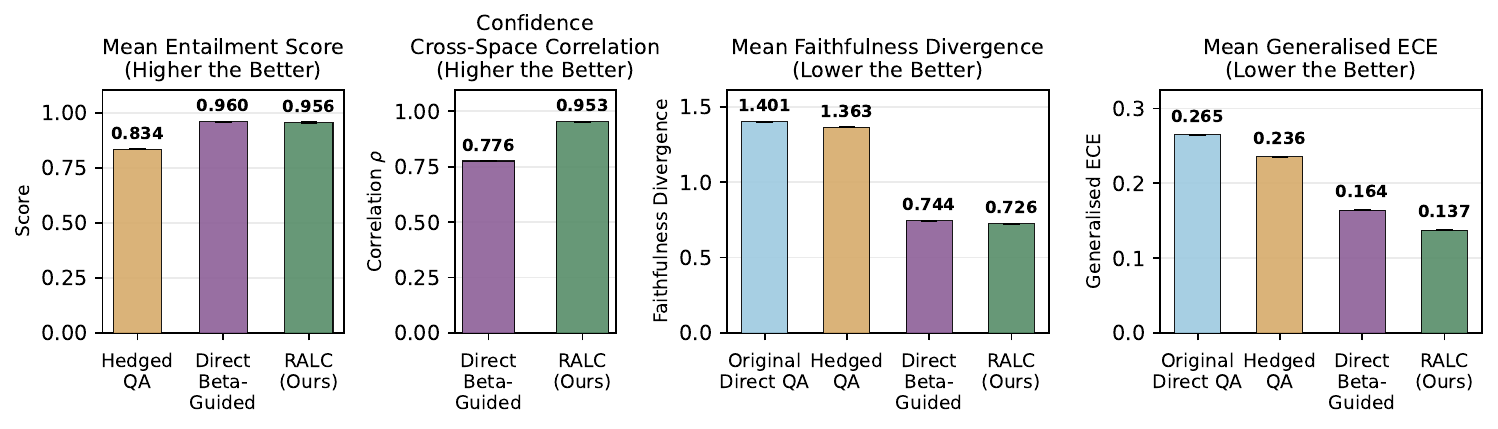}
\caption{Calibration effectiveness and quality comparison between RALC (averaged across all signals and models), Hedged QA, and Direct Beta-Guided Rewrite across content preservation (entailment), signal-to-language confidence correlation ($\rho$), linguistic-space Faithfulness Divergence, and linguistic-space generalised ECE. RALC matches Direct Beta-Guided Rewrite on content preservation, achieves a markedly higher signal-to-linguistic-space correlation, and outperforms both baselines on calibration and faithfulness.} 
    \label{fig:in-domain-calibration-direct-qa-to-hedged-qa-baseline}
\end{figure}

To contextualise RALC's linguistic calibration quality, we benchmark it against two baselines: \textbf{Hedged QA}, a black-box baseline that prompts the model to hedge without access to any calibrated signal \citep{yona2024can}, and \textbf{Direct Beta-Guided Rewrite}, a grey-box ablation of our pipeline in which the lexicon retrieval step is removed and the calibrated Beta distribution is passed directly to the rewriting model, relying on it to select appropriate hedging language without explicit linguistic grounding. This comparison isolates the contribution of structured lexicon retrieval over uncontrolled and partially controlled generation.

All performance metrics are measured in the linguistic space. We first apply our LLM ensemble to estimate the linguistic confidence expressed in the original uncalibrated Direct QA responses. For each calibration method, we then re-estimate linguistic confidence from the rewritten output and compute all metrics accordingly. Following \citet{gal2024semanticentropy}'s entailment setup, content preservation is scored as $p_{\mathrm{entail}} + 0.5 \cdot p_{\mathrm{neutral}}$ using DeBERTa-v3-Large-MNLI, and evaluated alongside signal-to-language confidence correlation ($\rho$), Faithfulness Divergence, and generalised ECE. As shown in Figure~\ref{fig:in-domain-calibration-direct-qa-to-hedged-qa-baseline}, RALC substantially outperforms Hedged QA across all metrics. Against Direct Beta-Guided Rewrite, RALC matches on content preservation whilst achieving lower Faithfulness Divergence and ECE, and attains a markedly higher Spearman $\rho$ between calibrated signal-space and re-estimated linguistic-space confidence, indicating that structured lexicon retrieval propagates the calibrated signal into language more reliably than unconstrained hedging selection.

\subsubsection{Cross-domain confidence calibration}

\begin{table}[t]
\centering
\caption{In-domain and cross-domain linguistic-space calibration metric percentage changes for both Faithfulness Divergence and generalised ECE. We report percentage change relative to the pre-calibration metrics (mean\,$\pm$\,std across models). Green text indicates calibration improvement (lower error). Semantic uncertainty is the strongest signal for RALC in improving both calibration and faithfulness across all three benchmarks in both in-domain and cross-domain settings.}
\label{tab:cross-domain-calibration}
\resizebox{\linewidth}{!}{
\begin{tabular}{p{2cm}p{2cm}lccc}
\toprule
\textbf{Metric} & \textbf{Signal} & \textbf{Train/Test} & {MMLU} & SQuAD~2.0 & {TruthfulQA} \\
\midrule
\multirow{9}{=}{\textbf{Faithfulness\\Divergence\\Mean\\Reduction}}
& \multirow{3}{=}{Linguistic\\Confidence}
& MMLU & \textcolor{green!70!black}{$\Delta$11.4$\pm$19.2\%} & \textcolor{green!70!black}{$\Delta$34.6$\pm$11.6\%} & \textcolor{green!70!black}{$\Delta$29.6$\pm$12.4\%} \\
& & SQuAD~2.0 & \textcolor{green!70!black}{$\Delta$42.7$\pm$4.2\%} & \textcolor{green!70!black}{$\Delta$62.0$\pm$2.3\%} & \textcolor{green!70!black}{$\Delta$59.8$\pm$3.9\%} \\
& & TruthfulQA & \textcolor{green!70!black}{$\Delta$27.0$\pm$17.2\%} & \textcolor{green!70!black}{$\Delta$65.8$\pm$2.2\%} & \textcolor{green!70!black}{$\Delta$60.5$\pm$4.6\%} \\
\cmidrule(lr){2-6}
& \multirow{3}{=}{Token\\Probability}
& MMLU & \textcolor{green!70!black}{$\Delta$10.2$\pm$23.2\%} & \textcolor{green!70!black}{$\Delta$52.6$\pm$8.1\%} & \textcolor{green!70!black}{$\Delta$56.4$\pm$6.2\%} \\
& & SQuAD~2.0 & \textcolor{green!70!black}{$\Delta$32.8$\pm$4.9\%} & \textcolor{green!70!black}{$\Delta$64.9$\pm$2.4\%} & \textcolor{green!70!black}{$\Delta$70.4$\pm$3.3\%} \\
& & TruthfulQA & \textcolor{green!70!black}{$\Delta$48.8$\pm$5.1\%} & \textcolor{green!70!black}{$\Delta$64.2$\pm$3.1\%} & \textcolor{green!70!black}{$\Delta$64.6$\pm$3.0\%} \\
\cmidrule(lr){2-6}
& \multirow{3}{=}{Semantic\\Uncertainty}
& MMLU & \textcolor{green!70!black}{$\Delta$21.6$\pm$13.9\%} & \textcolor{green!70!black}{$\Delta$36.1$\pm$12.4\%} & \textcolor{green!70!black}{$\Delta$43.0$\pm$9.8\%} \\
& & SQuAD~2.0 & \textcolor{green!70!black}{$\Delta$43.9$\pm$4.8\%} & \textcolor{green!70!black}{$\Delta$66.0$\pm$4.1\%} & \textcolor{green!70!black}{$\Delta$65.5$\pm$4.3\%} \\
& & TruthfulQA & \textcolor{green!70!black}{$\Delta$59.9$\pm$2.8\%} & \textcolor{green!70!black}{$\Delta$66.4$\pm$1.9\%} & \textcolor{green!70!black}{$\Delta$66.1$\pm$2.9\%} \\
\midrule
\multirow{9}{=}{\textbf{Generalised\\ECE\\Mean\\Reduction}}
& \multirow{3}{=}{Linguistic\\Confidence}
& MMLU & \textcolor{green!70!black}{$\Delta$38.1$\pm$8.0\%} & \textcolor{green!70!black}{$\Delta$20.7$\pm$7.6\%} & \textcolor{green!70!black}{$\Delta$12.2$\pm$6.4\%} \\
& & SQuAD~2.0 & \textcolor{green!70!black}{$\Delta$21.9$\pm$11.4\%} & \textcolor{green!70!black}{$\Delta$43.4$\pm$1.8\%} & \textcolor{green!70!black}{$\Delta$35.0$\pm$2.3\%} \\
& & TruthfulQA & \textcolor{green!70!black}{$\Delta$21.9$\pm$7.1\%} & \textcolor{green!70!black}{$\Delta$47.2$\pm$1.9\%} & \textcolor{green!70!black}{$\Delta$39.6$\pm$0.9\%} \\
\cmidrule(lr){2-6}
& \multirow{3}{=}{Token\\Probability}
& MMLU & \textcolor{green!70!black}{$\Delta$46.8$\pm$8.7\%} & \textcolor{green!70!black}{$\Delta$38.5$\pm$11.1\%} & \textcolor{green!70!black}{$\Delta$37.5$\pm$10.1\%} \\
& & SQuAD~2.0 & \textcolor{green!70!black}{$\Delta$23.5$\pm$2.2\%} & \textcolor{green!70!black}{$\Delta$50.6$\pm$1.0\%} & \textcolor{green!70!black}{$\Delta$61.4$\pm$1.1\%} \\
& & TruthfulQA & \textcolor{green!70!black}{$\Delta$60.2$\pm$2.6\%} & \textcolor{green!70!black}{$\Delta$48.6$\pm$2.4\%} & \textcolor{green!70!black}{$\Delta$42.3$\pm$2.3\%} \\
\cmidrule(lr){2-6}
& \multirow{3}{=}{Semantic\\Uncertainty}
& MMLU & \textcolor{green!70!black}{$\Delta$49.8$\pm$8.7\%} & \textcolor{green!70!black}{$\Delta$30.8$\pm$8.5\%} & \textcolor{green!70!black}{$\Delta$33.7$\pm$5.7\%} \\
& & SQuAD~2.0 & \textcolor{green!70!black}{$\Delta$30.1$\pm$16.9\%} & \textcolor{green!70!black}{$\Delta$58.7$\pm$3.8\%} & \textcolor{green!70!black}{$\Delta$49.8$\pm$4.1\%} \\
& & TruthfulQA & \textcolor{green!70!black}{$\Delta$62.3$\pm$3.1\%} & \textcolor{green!70!black}{$\Delta$56.9$\pm$4.0\%} & \textcolor{green!70!black}{$\Delta$45.0$\pm$3.1\%} \\
\bottomrule
\end{tabular}
}
\end{table}

Having established strong in-domain performance, we examine the cross-domain transferability of confidence signals through the lens of RALC. The assessment is independent of the performance of RALC; rather, it uses RALC as a diagnostic framework to evaluate signal stability under domain shift, specifically the reliability of each confidence signal as an input to the pipeline when the Platt scaling map trained on one dataset is applied to a different domain without retraining.

We train the Platt scaling calibration map on each of the three datasets in turn and evaluate the resulting RALC pipeline on all three datasets without retraining, yielding both in-domain (diagonal entries) and cross-domain (non-diagonal entries) conditions. We report pre-to-post-RALC percentage reductions in linguistic-space Faithfulness Divergence and generalised ECE \citep{wang2025calibratingexpressionscertainty} in Table~\ref{tab:cross-domain-calibration}. Raw metric value changes are provided in Appendix~\ref{appendix:cross-domain-calibration-value-chg}.

All three signals yield improvements across both metrics and all domain pairs. Semantic uncertainty exhibits the strongest cross-domain transferability, producing the highest gains with the lowest variance across models in both in-domain and cross-domain settings. As a result, the empirical evidence supports semantic uncertainty as the most robust confidence signal for RALC. 

In contrast, cross-domain calibrators occasionally outperform in-domain ones. This anomaly occurs when the target domain has a weak miscalibration bias, providing insufficient signal for its in-domain calibrator to learn a reliable correction. A cross-domain source with a stronger, more consistent bias learns a more decisive correction that transfers to the target domain, provided both share the same direction of miscalibration. We provide a detailed investigation in Appendix~\ref{sec:cross_domain_investigation}.

\section{Discussion and conclusion}
\label{section:discussion}
\paragraph{Limitations and future work}
As a downstream framework, RALC's calibration quality is ultimately bounded by the quality of the upstream confidence signal. Whilst semantic uncertainty is the most robust signal evaluated, its reliance on multi-round self-consistency sampling incurs significant inference cost. Identifying signals that match its performance at a lower computational expense is therefore an important direction for future work. Additionally, the hedging expression lexicon covers common confidence expressions and is not intended to represent the full landscape of linguistic uncertainty cues. Future work could investigate domain-specific hedging vocabularies and audience-adapted confidence scoring, which would enhance the specialisation and expressiveness of RALC in targeted deployment settings.

\paragraph{Conclusion}
In this work, we introduce a distributional treatment of linguistic confidence as a Beta distribution, define faithfulness as a complementary dimension of confidence evaluation, and present Faithfulness Divergence to quantify it from an information-theoretic perspective. Building on these foundations, we propose Retrieval-Augmented Linguistic Calibration (RALC), a principled post-hoc pipeline that calibrates confidence in the linguistic space, yielding well-calibrated and faithful responses that consistently outperform prompt-based baselines across models and benchmarks.

\clearpage
\bibliographystyle{unsrtnat}
\bibliography{main}

\clearpage
\appendix

\section{Theory}
\label{appendix:theory}
\subsection{Beta distribution estimation}
\label{appendix:beta-dist-estimation}

The Beta distribution is a natural choice for modelling random variables supported on $[0,1]$,
such as confidence scores and empirical accuracies.
A Beta distribution with parameters $(\alpha,\beta)$ has probability density function
\[
p(x \mid \alpha,\beta)
=
\frac{1}{\mathrm{B}(\alpha,\beta)} x^{\alpha-1} (1-x)^{\beta-1},
\quad x \in [0,1],
\]
where $\alpha>0$, $\beta>0$, and $\mathrm{B}(\alpha,\beta)$ denotes the Beta function.
Its mean, variance, and concentration factor are given by
\[
\mathbb{E}[X] = \frac{\alpha}{\alpha+\beta},
\qquad
\mathrm{Var}(X) = \frac{\alpha\beta}{(\alpha+\beta)^2(\alpha+\beta+1)},
\qquad
\kappa = \alpha+\beta.
\]

\subsubsection{Method of moments}
\label{appendix:beta-method-of-moments}
Let $\mathcal{S}_Z = \{s_t\}_{t=1}^{T_Z}$ denote the pseudo-observation set for a fixed target $Z$,
with $s_t \in [0,1]$.
Denote the empirical mean and variance by
\[
\bar{s}_Z = \frac{1}{T_Z} \sum_{t=1}^{T_Z} s_t,
\qquad
v_Z = \frac{1}{T_Z-1} \sum_{t=1}^{T_Z} (s_t - \bar{s}_Z)^2.
\]
The method of moments estimates $(\alpha,\beta)$ by matching these empirical moments
to the theoretical mean and variance of the Beta distribution.
Solving the resulting system yields
\[
\hat{\alpha}_Z
=
\bar{s}_Z
\left(
\frac{\bar{s}_Z(1-\bar{s}_Z)}{v_Z} - 1
\right),
\qquad
\hat{\beta}_Z
=
(1-\bar{s}_Z)
\left(
\frac{\bar{s}_Z(1-\bar{s}_Z)}{v_Z} - 1
\right),
\]
provided that $v_Z > 0$ and $v_Z < \bar{s}_Z(1-\bar{s}_Z)$.
When this condition is not met, two fallback cases are distinguished.

\textbf{Boundary-degenerate case} ($\bar{s}_Z \in \{0,1\}$, i.e.\ all observations are 0 or all are 1):
the mean-preserving concentration formula cannot be applied since one parameter would be zero.
Instead we set
\[
\hat{\alpha}_Z = \bar{s}_Z \cdot \kappa, \qquad \hat{\beta}_Z = (1 - \bar{s}_Z) \cdot \kappa,
\]
with $\kappa = T_Z$, and rely on the clipping step below to lift any zero parameter to $10^{-6}$.
This yields Beta($T_Z$, $10^{-6}$) when all observations are 1, placing nearly all mass near 1,
and Beta($10^{-6}$, $T_Z$) when all observations are 0, placing nearly all mass near 0.

\textbf{Interior-degenerate case} ($v_Z = 0$ or $v_Z \geq \bar{s}_Z(1-\bar{s}_Z)$
with $\bar{s}_Z \in (0,1)$): the observations are constant or insufficiently dispersed at an
interior value. We preserve the empirical mean by setting
\[
\hat{\alpha}_Z = \bar{s}_Z \cdot \kappa, \qquad \hat{\beta}_Z = (1 - \bar{s}_Z) \cdot \kappa,
\]
with $\kappa = T_Z$. This produces a Beta distribution with mean $\bar{s}_Z$ and high concentration
proportional to the number of observations, reflecting the certainty implied by the consistency
of the pseudo-observations.

In all cases, both parameters are clipped to a minimum of $10^{-6}$ to ensure numerical stability:
\[
\hat{\alpha}_Z \leftarrow \max(\hat{\alpha}_Z,\, 10^{-6}),
\qquad
\hat{\beta}_Z \leftarrow \max(\hat{\beta}_Z,\, 10^{-6}).
\]

\subsection{Classical instance-level calibration metrics with distribution generalisation}
\label{appendix:distributional-brier-nll}
\subsubsection{Expected Brier Score}

Let $S = \mathrm{Beta}(\alpha, \beta)$ denote the predictive confidence distribution, let $p \sim S$ denote the scalar confidence value drawn from it, and let $y \in \{0,1\}$ be the ground-truth label. The Expected Brier Score is defined as
\[
\mathbb{E}_{p \sim S}\left[(p - y)^2\right].
\]

This admits a closed form:
\[
\mathbb{E}[(p - y)^2]
=
\mathrm{Var}(p) + \left(\mathbb{E}[p] - y\right)^2,
\]
where
\[
\mathbb{E}[p] = \frac{\alpha}{\alpha + \beta}, 
\quad
\mathrm{Var}(p) = \frac{\alpha \beta}{(\alpha + \beta)^2 (\alpha + \beta + 1)}.
\]

Hence,
\[
\mathbb{E}[(p - y)^2]
=
\frac{\alpha \beta}{(\alpha + \beta)^2 (\alpha + \beta + 1)}
+
\left(\frac{\alpha}{\alpha + \beta} - y\right)^2.
\]

\subsection{Expected Negative Log-Likelihood (NLL)}

Given $S = \mathrm{Beta}(\alpha, \beta)$, $p \sim S$, and label $y \in \{0,1\}$, we define the distributional Negative Log-Likelihood as the expected Bernoulli log-loss under $S$:
\[
\mathcal{L}_{\mathrm{NLL}}(S, y)
=
\mathbb{E}_{p \sim S}\left[-\log p(y \mid p)\right],
\]
where $p(y \mid p) = p^y (1-p)^{1-y}$.

This yields the closed form:
\[
\mathcal{L}_{\mathrm{NLL}}(S, y)
=
\begin{cases}
\psi(\alpha + \beta) - \psi(\alpha), & y = 1, \\
\psi(\alpha + \beta) - \psi(\beta), & y = 0,
\end{cases}
\]
where $\psi(\cdot)$ denotes the digamma function.

Equivalently, this can be written as
\[
\mathcal{L}_{\mathrm{NLL}}(S, y)
=
y \left[\psi(\alpha + \beta) - \psi(\alpha)\right]
+
(1-y)\left[\psi(\alpha + \beta) - \psi(\beta)\right].
\]

\clearpage
\section{Further discussion on faithfulness}
\label{appendix:faithfulness-discussion}

\subsection{Distributional representation of linguistic confidence}
We represent response-level linguistic confidence as $S = \mathrm{Beta}(\alpha,\beta)$, where the mean $\mu = \alpha/(\alpha+\beta)$ captures the consensus perceived confidence across readers and the concentration $\kappa = \alpha+\beta$ captures the strength of that consensus. Two responses may elicit identical mean confidence yet differ substantially in concentration: one may induce consistent perceptions across readers whilst the other induces highly variable ones. Scalar representations discard this distinction; the distributional representation preserves it, and it is precisely this information that faithfulness evaluation requires.

\subsection{Faithfulness Divergence}
Faithfulness measures the degree of surprise induced by truth revelation. A response that elicits high mean confidence with high concentration is highly surprising when incorrect, as it represents a strongly held prior belief that the ground truth contradicts. The same misalignment expressed with low concentration is less surprising, since the prior was weakly held and requires only a modest update. Faithfulness Divergence (FD) operationalises this intuition.

Formally, for instance $i$ with correctness label $y_i$, the estimated confidence distribution serves as the prior $S_i = \mathrm{Beta}(\alpha_i, \beta_i)$. Upon observing $y_i$, the prior is updated by a single Bernoulli observation to yield the posterior $S_i^* = \mathrm{Beta}(\alpha_i + y_i,\, \beta_i + 1 - y_i)$. The KL divergence $\mathrm{KL}(S_i^* \| S_i)$ quantifies the normalised magnitude of the required belief revision. However, KL divergence alone does not account for the strength of agreement underlying the prior. Under the Beta--Bernoulli model, the concentration $\alpha_i + \beta_i$ is interpretable as the effective sample size of the prior: a larger concentration encodes a more strongly held belief, and an identical KL divergence therefore represents a larger total epistemic adjustment when the prior is more concentrated \citep{morita2007determining}. FD scales the KL divergence by this effective sample size,
\[
\mathrm{FD}_i := (\alpha_i + \beta_i)\cdot\mathrm{KL}\!\left(S_i^* \,\|\, S_i\right),
\]
yielding a scalar that quantifies the degree of surprise induced by truth revelation, scaled by the strength of agreement encoded in the prior.

\subsection{Scope and intended use}
FD is designed as a diagnostic measurement tool, not as a proper scoring rule or a training objective. It quantifies the surprise induced at the instance level when the ground truth is revealed, providing a complementary lens on confidence quality that population-level metrics such as ECE do not capture. The goal of a well-calibrated response is not to minimise FD in isolation, but to express confidence that is consistent with the model's actual uncertainty; FD measures how far a given response falls from that standard. It should therefore be interpreted alongside calibration metrics rather than treated as a sole optimisation target.

\clearpage
\subsection{Additional Faithfulness Divergence ablation studies}
\label{appendix:additional-fd-ablations}

In addition to the empirical ablation in Table~\ref{tab:fd_miscalibration}, we conduct
theoretical ablations to further illustrate the unique properties of Faithfulness Divergence and its alignment with our definition of surprise upon truth revelation, compared against alternative metrics including KL divergence, expected Brier score, and
expected NLL. Two controlled settings are examined: varying the confidence
mean under a fixed concentration, and varying concentration under a fixed mean, each
evaluated against a binary ground-truth label. 

\begin{figure}[h!]
    \centering
    \includegraphics[width=0.85\linewidth]{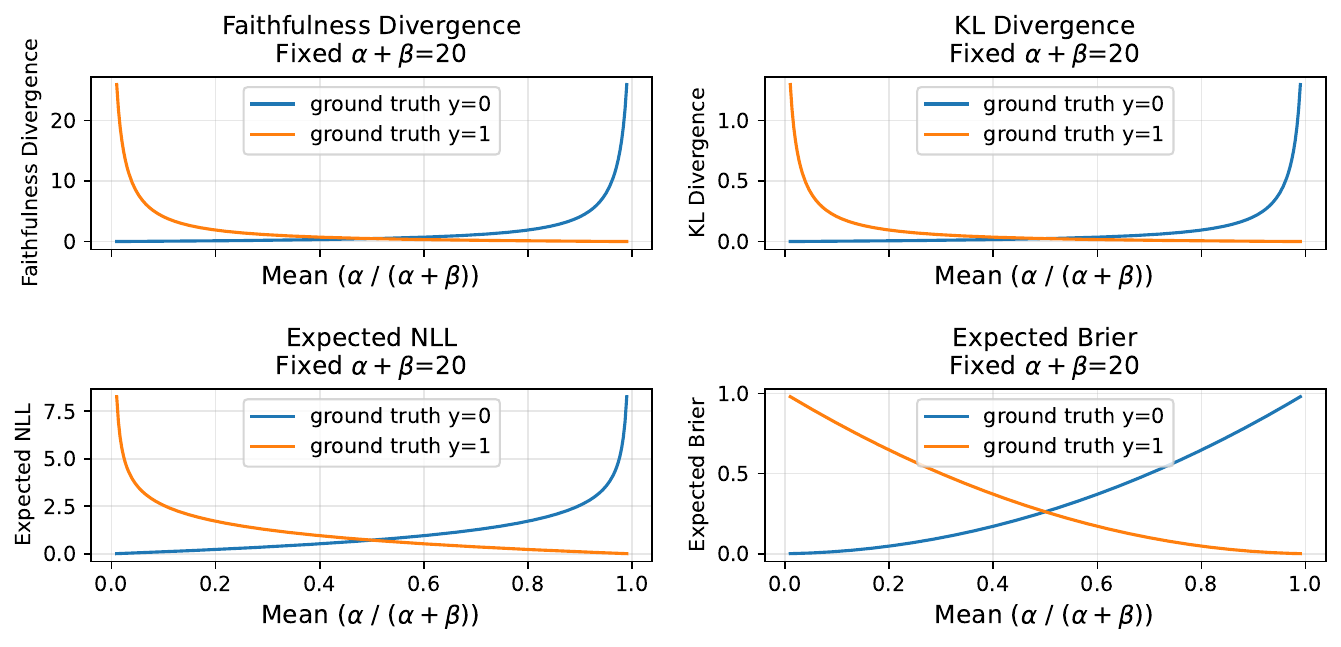}
    \caption{Faithfulness Divergence, KL divergence, expected Brier score, and expected NLL
    under varying confidence means with fixed concentration ($\alpha + \beta = 20$) against
    a binary ground-truth label. All metrics increase monotonically as the mean deviates from the
    ground-truth label, reflecting greater surprise upon truth revelation for more
    misaligned beliefs.}
    \label{fig:fd_ablation_theoretical_fixed_conc}
\end{figure}

\begin{figure}[h!]
    \centering
    \includegraphics[width=0.85\linewidth]{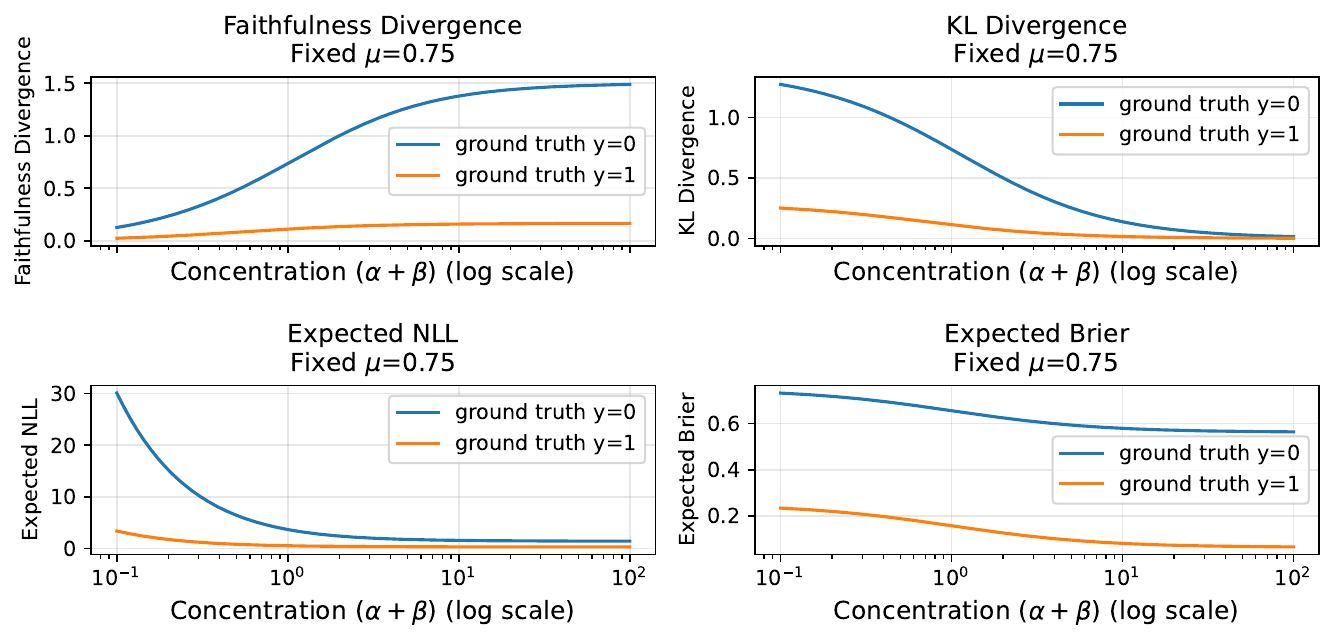}
    \caption{Faithfulness Divergence, KL divergence, expected Brier score, and expected NLL
    under varying concentration with a fixed misaligned mean ($\mu = 0.75$) against a
    binary ground-truth label. Only Faithfulness Divergence correctly increases monotonically with concentration, encoding the
    intuition that a strongly held incorrect belief induces greater surprise upon truth
    revelation than a weakly held one of equal mean.}
    \label{fig:fd_ablation_theoretical_fixed_mean}
\end{figure}

Figure~\ref{fig:fd_ablation_theoretical_fixed_conc} shows that all four metrics correctly
capture mean deviation from the ground-truth label as increasing surprise, assigning
monotonically higher values as the confidence mean moves further from the outcome. However, Figure~\ref{fig:fd_ablation_theoretical_fixed_mean} demonstrates that only FD encodes concentration as an amplifier of misalignment, assigning monotonically higher surprise as concentration increases for a fixed misaligned mean, whilst the
alternative metrics behave otherwise. FD therefore uniquely quantifies surprise upon truth revelation in accordance with our definition, which requires that a more strongly held incorrect belief be treated as more surprising.

\clearpage
\section{Confidence evaluation implementation}
\label{appendix:conf-eval-details}
\subsection{QA prompts}
\label{appendix:qa-prompts}
We detail the exact prompts used to sample responses across MMLU, SQuAD~2.0, and TruthfulQA with different confidence estimation methods, including linguistic confidence, token probability, and semantic uncertainty.

\subsubsection{MMLU}
\begin{tcolorbox}[colback=white, colframe=black,
title=\textbf{Direct QA prompt}, fonttitle=\bfseries, boxrule=0.8pt, fontupper=\scriptsize, parbox=false]
Answer the following question using a succinct (at most one sentence) and full answer.

\{ question \}

\{\% for choice in choices \%\}

\{ 'ABCDEFGHIJKLMNOPQRSTUVWXYZ'[loop.index0] \}. \{ choice \}

\{\% endfor \%\}

Answer: [Return the answer letter only]
\end{tcolorbox}

\begin{tcolorbox}[colback=white, colframe=black,
title=\textbf{Hedged QA prompt}, fonttitle=\bfseries, boxrule=0.8pt, fontupper=\scriptsize, parbox=false]
Answer the following question using a succinct (at most one sentence) and full answer.
If you are uncertain about your answer to the question,
convey this uncertainty linguistically by precisely hedging this answer. Please include the option letter in your answer.

Question: \{ question \}

\{\% for choice in choices \%\}

\{ 'ABCDEFGHIJKLMNOPQRSTUVWXYZ'[loop.index0] \}. \{ choice \}

\{\% endfor \%\}

Answer:
\end{tcolorbox}

\subsubsection{SQuAD~2.0}
\begin{tcolorbox}[colback=white, colframe=black,
title=\textbf{Direct QA prompt}, fonttitle=\bfseries, boxrule=0.8pt, fontupper=\scriptsize, parbox=false]
Answer the following question using a succinct (at most one sentence) and full answer.

Title: \{ title \}

Background: \{ context \}

Question: \{ question \}

Answer:
\end{tcolorbox}

\begin{tcolorbox}[colback=white, colframe=black,
title=\textbf{Hedged QA prompt}, fonttitle=\bfseries, boxrule=0.8pt, fontupper=\scriptsize, parbox=false]
Answer the following question using a succinct (at most one sentence) and full answer.
If you are uncertain about your answer to the question,
convey this uncertainty linguistically by precisely hedging this answer.

Title: \{ title \}

Background: \{ context \}

Question: \{ question \}

Answer:
\end{tcolorbox}

\subsubsection{TruthfulQA}
\begin{tcolorbox}[colback=white, colframe=black,
title=\textbf{Direct QA prompt}, fonttitle=\bfseries, boxrule=0.8pt, fontupper=\scriptsize, parbox=false]
Answer the following question using a succinct (at most one sentence) and full answer.

Question: \{question\}

Answer:
\end{tcolorbox}

\begin{tcolorbox}[colback=white, colframe=black,
title=\textbf{Hedged QA prompt}, fonttitle=\bfseries, boxrule=0.8pt, fontupper=\scriptsize, parbox=false]
Answer the following question using a succinct (at most one sentence) and full answer.
If you are uncertain about your answer to the question,
convey this uncertainty linguistically by precisely hedging this answer.

Question: \{question\}

Answer:
\end{tcolorbox}

\clearpage
\subsection{LLM linguistic confidence evaluator}
\subsubsection{Evaluator prompt}
An LLM linguistic confidence evaluator ensemble is prompted as follows. We use an ensemble of models rather than a single evaluator to capture the complex linguistic relationships arising from co-occurring cues and their contextual interactions within a statement, whilst averaging out idiosyncratic biases of individual models. To align with human perception, we provide the LLMs with human-annotated linguistic cues and their associated confidence profiles from \citet{tao2025largelanguagemodelsexpress} as reference. The LLM is then asked to return a confidence score between 0 and 100 based solely on the linguistic cues present in the sentence, without using any external or prior knowledge to assess the knowledge conveyed by the sentence. The extracted output score is then normalised to $[0,1]$. 

\begin{tcolorbox}[colback=white, colframe=black,  title=\textbf{LLM linguistic confidence evaluator prompt}, fonttitle=\bfseries, boxrule=0.8pt, fontupper=\scriptsize, parbox=false]

Please provide only a confidence score between 0 and 100, based solely on the degree of confidence expressed in the tone and linguistic cues of the following sentence, without using any external or prior knowledge to assess the knowledge conveyed by the sentence. 
Please pay attention to the hedging language used and the overall assertiveness of the statement. 
If the sentence abstains from answering by pointing out the insufficiency of information with a firm tone, the score should be towards 100. 
If the sentence contains random guesses or abstention, the score should be towards 0. If the sentence is stated with strong certainty or no hedging, the score should be towards 100. 
If the sentence does not contain any hedging language or is a succinct, decisive short answer, the score should be towards 100, too.

To align with human perception, here are some human-annotated linguistic cues with perceived confidence profiles (mean and standard deviation) for your reference: 
\{human\_annotated\_cues\}

Here is the sentence:
\{sentence\}

Confidence Score: [Return only a number between 0 and 100 without any additional text or explanation]
\end{tcolorbox}

\subsubsection{LLM ensemble vs. human benchmark}
\label{appendix:llm-ensemble-human-benchmark}
\begin{figure}[h!]
    \centering
    \includegraphics[width=0.7\linewidth]{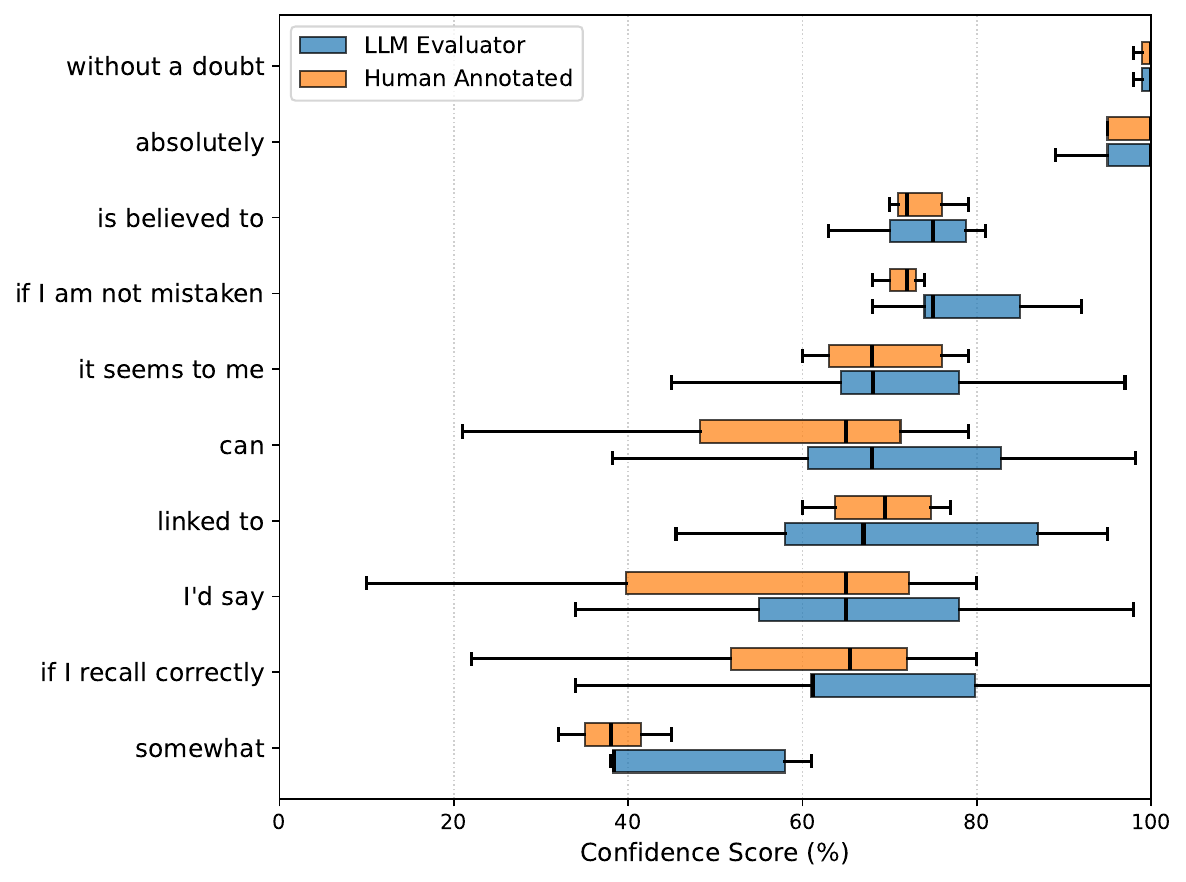}
    \caption{LLM vs. human perceived linguistic confidence on the human-annotated benchmark of \citet{tao2025largelanguagemodelsexpress}. The LLM ensemble largely follows human confidence annotations across confidence levels.}
    \label{fig:llm-human-confidence-alignment}
\end{figure}

We compare the LLM-ensemble confidence scores against human annotations on the benchmark of \citet{tao2025largelanguagemodelsexpress}. The benchmark consists of human-annotated confidence scores across various statements. We employ our LLM ensemble to generate confidence scores for each statement in a similar manner to human annotators. The result in Figure~\ref{fig:llm-human-confidence-alignment} and Table~\ref{tab:llm-human-correlation} confirm that the ensemble largely matches human confidence judgements for common hedging expressions.

\begin{table}[h!]
    \centering
    \caption{Rank and linear correlations between LLM-ensemble and human-annotated confidence scores on the benchmark of \citet{tao2025largelanguagemodelsexpress}. All $p$-values are $< 10^{-10}$, rejecting the null hypothesis of zero correlation ($H_0\colon \rho = 0$) and confirming that the ensemble reliably reproduces human perception of linguistic confidence cues.}
    \label{tab:llm-human-correlation}
    \begin{tabular}{lcc}
        \toprule
        \textbf{Metric} & \textbf{Coefficient} & \textbf{$p$-value} \\
        \midrule
        Spearman $\rho$ & 0.8535 & $< 10^{-10}$ \\
        Pearson $r$     & 0.8450 & $< 10^{-10}$ \\
        Kendall $\tau$  & 0.6909 & $< 10^{-10}$ \\
        \bottomrule
    \end{tabular}
\end{table}

\subsection{Grader prompt}
\begin{tcolorbox}[colback=white, colframe=black,  title=\textbf{LLM grader prompt}, fonttitle=\bfseries, boxrule=0.8pt, fontupper=\scriptsize, parbox=false]

If the predicted answer matches, implies or covers the correct answer, the grade is CORRECT.

If the predicted answer does not match, imply or cover the correct answer, the grade is INCORRECT. Do NOT grade it as INCORRECT if the predicted answer abstains from answering (e.g. ``I don't know the answer...'' or ``I have no idea...'').

If the predicted answer is empty, none or abstention (e.g. ``I don't know the answer...'' or ``I have no idea...''), grade the predicted answer as NOT\_ATTEMPTED instead of CORRECT or INCORRECT. If the predicted answer makes an attempt (even random guesses), do not grade it as NOT\_ATTEMPTED.

Ignore any explanation or linguistic cues present in the predicted answer. Don't apologise or correct yourself if there was a mistake; we are just trying to grade the answer. 

Question: \{question\}

Correct answer: \{correct\_answer\}

Predicted answer: \{``'' if predicted\_answer is None else predicted\_answer\}

Grade the predicted answer of this new question as one of:

A: CORRECT

B: INCORRECT

C: NOT\_ATTEMPTED

Just return one of the letters ``A'', ``B'', or ``C'', with no text around it.
\end{tcolorbox}

\subsection{LLM-based semantic clustering prompt}
\label{appendix:semantic-clustering-prompt}
\begin{tcolorbox}[
  colback=white, 
  colframe=black,
  title=\textbf{LLM-based semantic clustering prompt}, 
  fonttitle=\bfseries, 
  boxrule=0.8pt, 
  fontupper=\scriptsize, parbox=false
]

You are a strict JSON generator. Group semantically equivalent candidate responses to the same question. Ignore any linguistic markers of uncertainty or hedging and focus solely on the core meaning of the responses. 

Return a JSON object with a single key \texttt{"semantic\_ids"}, a list of integers aligned with the response order. Responses that are semantically equivalent (bidirectional entailment) must share the same integer id. Use 0-based ids. Semantic ids represent the semantic cluster assignment for each response. Return ONLY the JSON object, no extra text.

For instance, given the question and candidate responses:

Question: What is the capital of France?

Candidate responses:\\
    0: `I guess Paris is the capital of France.'\\
    1: `Paris is the capital city of France.'\\
    2: `The capital of France is Berlin.'\\

The correct JSON output would be:

\texttt{\{"semantic\_ids": [0, 0, 1]\}}

Now, please group the following candidate responses to the given question and return the JSON object:

Question: \{question\}

Candidate responses: \{responses\}

\texttt{\{"semantic\_ids": [...]\}}

\end{tcolorbox}

\clearpage
\section{Post-hoc linguistic calibration implementation}
\label{appendix:linguistic-calibration-implementation}

\subsection{Lexicon construction}
\label{appendix:lexicon-construction}

The lexicon is constructed in three stages: hedging expression sourcing, confidence score collection, and Beta distribution fitting.

\paragraph{Hedging expression sourcing}
A curated set of $K$ hedging expressions spanning the full confidence spectrum is generated by prompting Claude-Sonnet-4.6 to produce words and phrases that humans use to convey varying degrees of certainty, from expressions of complete ignorance (e.g.\ ``I have no idea'', ``my random guess is'') to expressions of near-certain belief (e.g.\ ``without a doubt'', ``I can confirm''). 

\begin{tcolorbox}[colback=white, colframe=black,  title=\textbf{Hedging expression sourcing prompt}, fonttitle=\bfseries, boxrule=0.8pt, fontupper=\scriptsize, parbox=false]
Generate a Python list of words or expressions that humans use to convey the level of confidence, certainty, or hedging in their statements (without a subject, only the linguistic cues). These words should include common hedging phrases, adverbs, and qualifiers that indicate varying degrees of certainty or uncertainty, from extremely low confidence (like I do not know, my random guess is, etc) to high confidence (certain, sure, definitely). 

The list should be comprehensive and cover a wide range of expressions used in everyday language as well as in academic and professional contexts.
\end{tcolorbox}

\paragraph{Confidence score collection}
For each hedging expression $w_k$, GPT-OSS-20B \citep{agarwal2025gpt} generates 20 candidate sentences by rewriting a randomly selected non-verifiable statement drawn from a fixed pool of 12 non-verifiable template sentences, with instructions to incorporate $w_k$ naturally and to avoid introducing additional hedging cues. Each generated sentence is then independently evaluated by three LLM evaluators (Llama-3.1-8B-Instruct \citep{meta2024llama31}, Qwen3-8B \citep{qwen3}, Mistral-7B-Instruct-v0.3 \citep{mistralai2023mistral7b}), each prompted to assign a perceived-confidence score on a 0--100 scale based solely on the linguistic cues present, ignoring factual content. Human-annotated reference profiles from \citet{tao2025largelanguagemodelsexpress} are provided in-context to anchor model ratings to human perception. Each evaluator scores every sentence 3 times (temperature $= 1$), yielding up to $20 \times 3 \times 3 = 180$ raw scores per hedging expression.

\begin{tcolorbox}[colback=white, colframe=black,  title=\textbf{Non-verifiable template sentences}, fonttitle=\bfseries, boxrule=0.8pt, fontupper=\scriptsize, parbox=false]
"There is a correlation between X and Y."\\
"It rains tomorrow."\\
"The experiment shows a significant effect."\\
"The new policy improves the economy."\\
"The medication is effective in treating the disease."\\
"The new product is successful in the market."\\
"The neighbour is home."\\
"The movie is good."\\
"The restaurant serves delicious food."\\
"The city is the oldest in the country."\\
"The book is informative."\\
"The report is not accurate."
\end{tcolorbox}

\begin{tcolorbox}[colback=white, colframe=black,  title=\textbf{Non-verifiable sentence rewriting prompt}, fonttitle=\bfseries, boxrule=0.8pt, fontupper=\scriptsize, parbox=false]
Given a linguistic cue: "\{word\}", rewrite one of the following non-verifiable statements to naturally include this cue to convey the intended level of confidence, certainty, or hedging.
Please do not use other hedging words, hedging phrases or linguistic cues in the sentence other than the specified linguistic cue.

Example sentences to rewrite:
\{selected\_sentence\}

Do not use other hedging words or linguistic cues in the sentence. Do not combine linguistic cues. Do not include labels like "Example:" or "Sentence:". Just provide the statement.
\end{tcolorbox}

\paragraph{Beta distribution fitting}
All raw scores for a given expression are aggregated across sentences, evaluators, and repeated passes, then normalised to $[0,1]$ and clipped to $(10^{-6},\, 1 - 10^{-6})$ to avoid boundary degeneracy. A Beta distribution is fitted to the pooled scores by maximum likelihood estimation (with fixed support $[0,1]$), yielding the lexicon entry $\bigl(w_k,\, \mathrm{Beta}(\alpha_k, \beta_k)\bigr)$. The resulting lexicon $\{(w_k,\, \mathrm{Beta}(\alpha_k, \beta_k))\}_{k=1}^{K}$ is used at inference time for Wasserstein-distance-based retrieval.

\subsection{Sample hedging expressions from the lexicon}
\label{appendix:lexicon-samples}

\begin{figure}[h!]
    \centering
    \includegraphics[width=\linewidth]{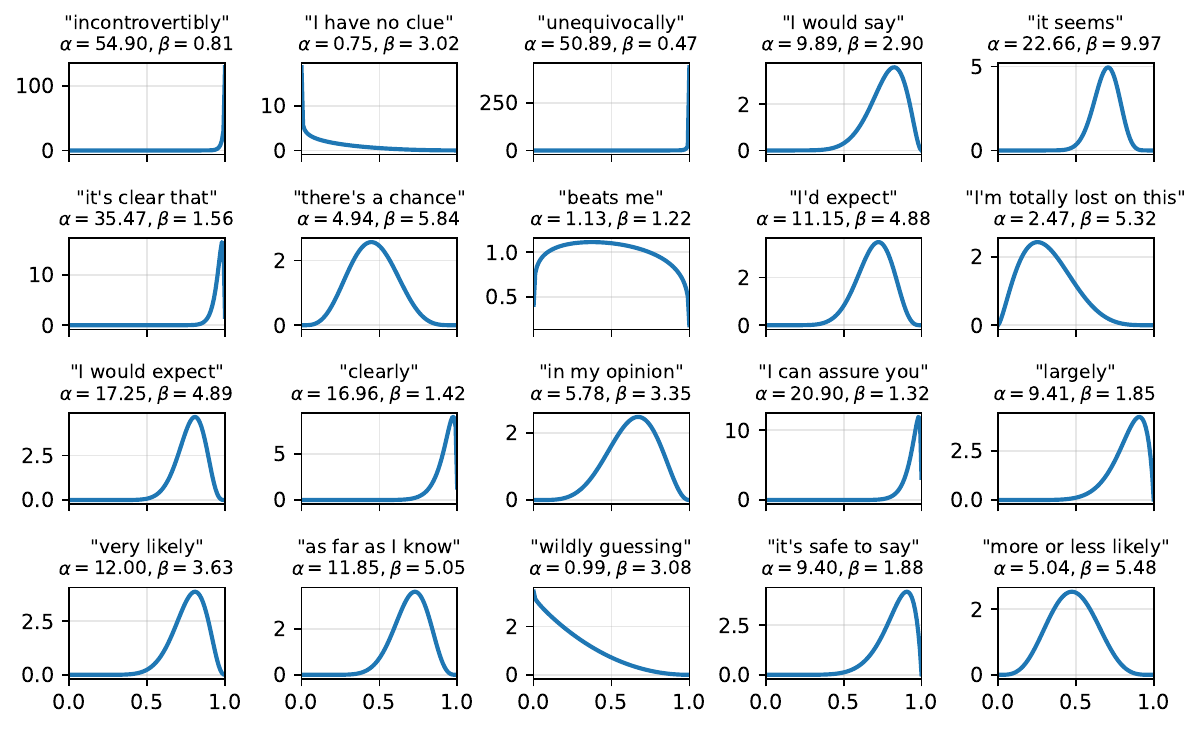}
    \caption{Sample hedging expressions from the lexicon, with their corresponding Beta distributions over perceived confidence by our LLM ensemble.}
    \label{fig:example-hedging-word-beta-distributions-lexicon}
\end{figure}

Figure~\ref{fig:example-hedging-word-beta-distributions-lexicon} shows sample hedging expressions from the lexicon, with their corresponding Beta distributions over perceived confidence by our LLM ensemble.

\subsection{Retrieval-augmented linguistic calibration rewriting prompt}

\label{appendix:retrieval-augmented-linguistic-calibration}
\begin{tcolorbox}[colback=white, colframe=black,  title=\textbf{Retrieval-augmented linguistic calibration prompt}, fonttitle=\bfseries, boxrule=0.8pt, fontupper=\scriptsize, parbox=false]

Given an original response and a list of target hedging words with their confidence profiles (Beta Distributions), rewrite the response to appropriately reflect the confidence level indicated by the set of target hedging words. 
You must preserve the original meaning of the response, as we are only adjusting the tone to match the confidence level suggested by the hedging words. Ensure the new response sounds natural and fluent. 

Original response: My answer to the question is: "\{response\}"

Target hedging words with confidence profiles: \{hedges\}

Please return only the rewritten sentence without any explanation.

New response: 
\end{tcolorbox}

\subsection{Direct Beta-guided rewriting calibration prompt}

\begin{tcolorbox}[colback=white, colframe=black,  title=\textbf{Direct Beta-guided rewriting prompt}, fonttitle=\bfseries, boxrule=0.8pt, fontupper=\scriptsize, parbox=false]

Given an original response and a Beta distribution, rewrite the response to appropriately reflect the confidence level indicated by the given Beta distribution by using hedging language. 
You must preserve the original meaning of the response, as we are only adjusting the tone to match the confidence level suggested by the hedging words. Ensure the new response sounds natural and fluent. 

Original response: My answer to the question is: "\{response\}"

Target Beta distribution: Beta(alpha=\{alpha:.2f\}, beta=\{beta:.2f\})

Please return only the rewritten sentence without any explanation.

New response: 

\end{tcolorbox}

\subsection{Choice of k for hedging expression retrieval}
\label{appendix:top-k-retrieval-ablation}
\begin{figure}[h!]
    \centering
    \includegraphics[width=0.85\linewidth]{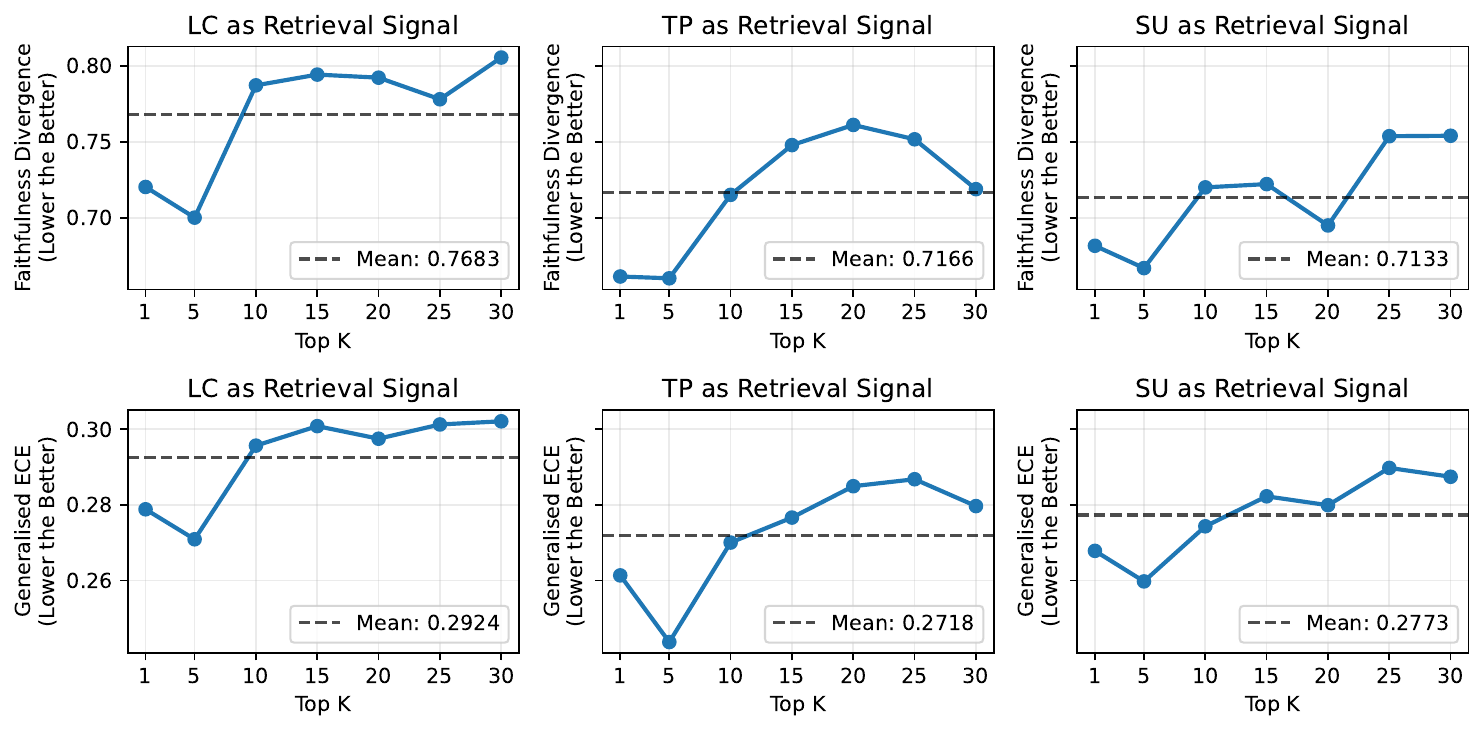}
    \caption{Impact of the choice of $k$ for the KNN retrieval of hedging expressions in RALC pipeline on Faithfulness Divergence and generalised ECE across different confidence signals for Llama-3.1-8B-Instruct on the TruthfulQA dataset. The results show that both metrics are not highly sensitive to the choice of $k$ within a reasonable range, with $k=5$ showing consistently better marginal performance in the exploration range.}
    \label{fig:top_k-ablation-results}
\end{figure}

Given the pre-constructed lexicon of hedging expressions, we perform an ablation study on the choice of $k$ for the KNN retrieval of hedging expressions in the RALC pipeline. The following figure shows the impact on Faithfulness Divergence and generalised ECE for different choices of $k$ across linguistic confidence (LC), token probability (TP), and semantic uncertainty (SU) as retrieval signals for Llama-3.1-8B-Instruct on the TruthfulQA dataset. Figure~\ref{fig:top_k-ablation-results} shows that both metrics are not highly sensitive to the choice of $k$, with $k=5$ showing consistently better marginal performance in the exploration landscape with the lowest Faithfulness Divergence and generalised ECE. Therefore, we choose $k=5$ for both the in-domain and cross-domain calibration experiments in this work.

\subsection{Confidence distribution profiles across different estimators}
\begin{figure}[h!]
    \centering
    \includegraphics[width=\linewidth]{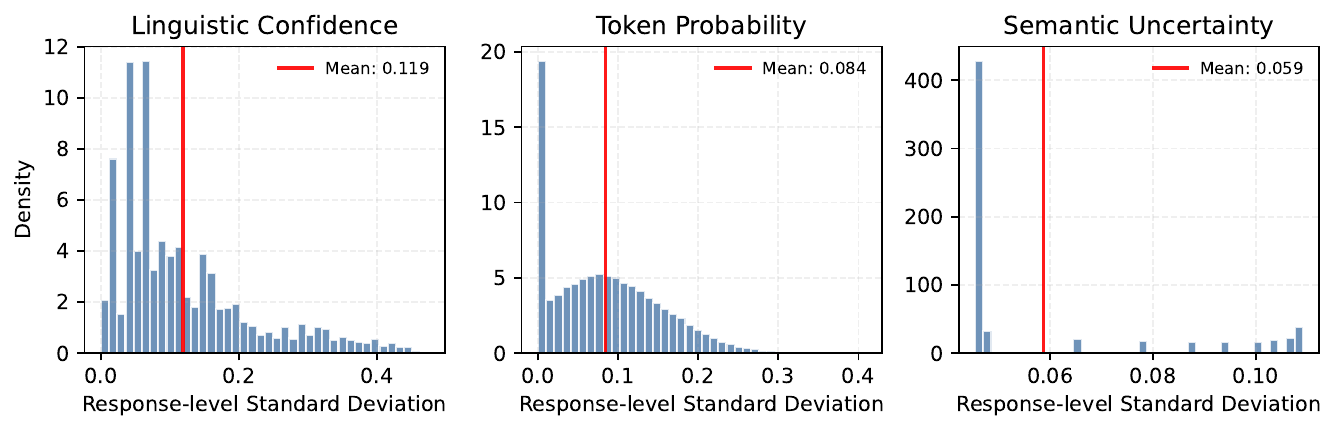}
    \caption{For each confidence signal for Direct QA responses, we plot the distribution of the standard deviation of the confidence distribution across responses.}
    \label{fig:confidence-distribution-profiles-across-estimators}
\end{figure}

Figure~\ref{fig:confidence-distribution-profiles-across-estimators} shows the distribution of confidence standard deviations across responses for each signal. Linguistic confidence exhibits the highest variability, whilst token probability and semantic uncertainty each produce a substantial proportion of zero-variance distributions, arising when responses share identical token probability profiles or collapse into a single semantic cluster. These degenerate cases are handled by clipping the $(\alpha, \beta)$ parameters as specified in Appendix~\ref{appendix:beta-method-of-moments}, ensuring compatibility with both the calibration map and the Wasserstein-based retrieval step.

\clearpage
\subsection{Calibration map ablation study}
\label{appendix:calibration-map-comparison}

\begin{table}[h!]
\centering
\caption{Ablation over signal-space calibration maps, averaged across all datasets and models ($\pm 1$ standard deviation). Bold green denotes the best (lowest) value per column. Platt scaling achieves the lowest error in five of six columns and is adopted in the RALC pipeline.}
\label{tab:calibration-map-ablation}
\resizebox{\linewidth}{!}{%
\begin{tabular}{lcccccc}
\toprule
& \multicolumn{2}{c}{\textbf{Linguistic confidence}} & \multicolumn{2}{c}{\textbf{Token probability}} & \multicolumn{2}{c}{\textbf{Semantic uncertainty}} \\
\cmidrule(lr){2-3}\cmidrule(lr){4-5}\cmidrule(lr){6-7}
\textbf{Method} & Gen. ECE & FD & Gen. ECE & FD & Gen. ECE & FD \\
\midrule
Uncalibrated
  & $0.280_{\pm 0.107}$
  & $1.487_{\pm 0.628}$
  & $0.262_{\pm 0.111}$
  & $68.879_{\pm 96.520}$
  & $0.267_{\pm 0.114}$
  & $2.112_{\pm 0.762}$ \\
Platt scaling \citep{PlattScaling1999}
  & $\mathbf{\textcolor{green!70!black}{0.109}}_{\pm 0.018}$
  & $\mathbf{\textcolor{green!70!black}{0.486}}_{\pm 0.042}$
  & $\mathbf{\textcolor{green!70!black}{0.085}}_{\pm 0.032}$
  & $\mathbf{\textcolor{green!70!black}{0.500}}_{\pm 0.060}$
  & $\mathbf{\textcolor{green!70!black}{0.060}}_{\pm 0.014}$
  & $0.506_{\pm 0.051}$ \\
Isotonic regression \citep{Zadrozny-isotonic-reg-2002}
  & $0.114_{\pm 0.019}$
  & $0.685_{\pm 0.530}$
  & $0.090_{\pm 0.034}$
  & $1.646_{\pm 1.495}$
  & $0.065_{\pm 0.017}$
  & $0.718_{\pm 0.439}$ \\
Histogram binning \citep{Zadrozny-histogram-binning-2001}
  & $0.110_{\pm 0.019}$
  & $0.488_{\pm 0.042}$
  & $0.090_{\pm 0.039}$
  & $0.514_{\pm 0.104}$
  & $0.060_{\pm 0.017}$
  & $\mathbf{\textcolor{green!70!black}{0.502}}_{\pm 0.047}$ \\
Temperature scaling \citep{guo2017calibrationmodernneuralnetworks}
  & $0.127_{\pm 0.022}$
  & $0.547_{\pm 0.148}$
  & $0.102_{\pm 0.032}$
  & $0.646_{\pm 0.380}$
  & $0.080_{\pm 0.031}$
  & $0.520_{\pm 0.056}$ \\
\bottomrule
\end{tabular}%
}
\end{table}

We ablate the signal-space calibration map across Platt scaling \citep{PlattScaling1999}, isotonic regression \citep{Zadrozny-isotonic-reg-2002}, histogram binning \citep{Zadrozny-histogram-binning-2001}, and temperature scaling \citep{guo2017calibrationmodernneuralnetworks}, applied to the distribution means of all three confidence signals and evaluated on generalised ECE and Faithfulness Divergence in the signal space. The results are averaged across all datasets and models and reported in Table~\ref{tab:calibration-map-ablation}. Platt scaling is the best performer in the signal space across both metrics and all confidence signals, whilst isotonic regression exhibits instability on small calibration sets, histogram binning trails on FD despite being competitive on ECE, and temperature scaling performs worst overall. We therefore adopt Platt scaling as the signal-space calibration map in the RALC pipeline.

\subsection{Confidence signal propagation quality}
We evaluate the quality of RALC by measuring the correlation between the calibrated confidence signal and the linguistic confidence in the rewritten responses perceived by our LLM evaluator ensemble for each confidence signal. Our pipeline accurately propagates the calibrated confidence signal into language, as evidenced by a positive Spearman's correlation $\rho$ consistently above 0.9 across all confidence signals. 

\begin{figure}[h!]
    \centering
    \includegraphics[width=\linewidth]{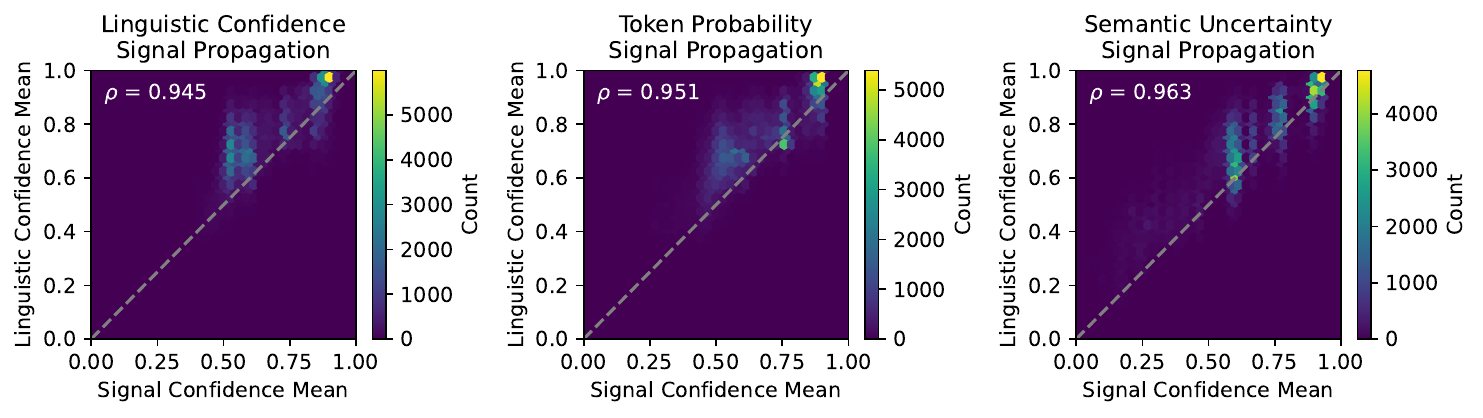}
    \caption{We evaluate the quality of RALC by measuring the correlation between the calibrated confidence signal and the linguistic confidence in the rewritten responses perceived by our LLM evaluator ensemble. Across all confidence signals, our pipeline effectively propagates the calibrated confidence signal into language, as evidenced by a positive correlation consistently well above 0.9.}
    \label{fig:direct-qa-confidence-signal-propagation-quality}
\end{figure}

\clearpage
\section{Additional results}
\label{appendix:additional-results}

\subsection{Additional performance metrics}

In addition to faithfulness and calibration, we assess discriminative performance using AUROC, computed on the means of each confidence distribution, averaged across all five models. An AUROC above 0.5 indicates that higher confidence tends to correlate with correctness, with 1.0 being perfect discrimination; a value near 0.5 reflects a signal no better than chance. Table~\ref{tab:accuracy-auroc} reports the pre-calibration signal profile where semantic uncertainty and token probability achieve stronger discrimination than linguistic confidence across all datasets, motivating the use of alternative confidence signals in our RALC pipeline beyond linguistic confidence.

\label{appendix:additional-performance-metrics}
\begin{table}[h!]
\centering
\caption{Model accuracy, mean confidence, and AUROC of each confidence signal mean across datasets, averaged across all five models.}
\label{tab:accuracy-auroc}
\small
\begin{tabular}{llccc}
\toprule
\textbf{Dataset} & \textbf{Signal} & \textbf{Acc.} & \textbf{Mean Conf.} & \textbf{AUROC} \\
\midrule
\multirow{3}{*}{MMLU}
  & Ling.\ Confidence & \multirow{3}{*}{$0.719 \pm 0.120$} & $0.784 \pm 0.014$ & $0.538 \pm 0.026$ \\
  & Token Probability &                                      & $0.936 \pm 0.045$ & $0.609 \pm 0.075$ \\
  & Sem.\ Uncertainty &                                      & $0.891 \pm 0.054$ & $0.652 \pm 0.076$ \\
\cmidrule(lr){1-5}
\multirow{3}{*}{SQuAD~2.0}
  & Ling.\ Confidence & \multirow{3}{*}{$0.564 \pm 0.100$} & $0.840 \pm 0.015$ & $0.497 \pm 0.029$ \\
  & Token Probability &                                      & $0.838 \pm 0.104$ & $0.647 \pm 0.035$ \\
  & Sem.\ Uncertainty &                                      & $0.884 \pm 0.038$ & $0.647 \pm 0.070$ \\
\cmidrule(lr){1-5}
\multirow{3}{*}{TruthfulQA}
  & Ling.\ Confidence & \multirow{3}{*}{$0.493 \pm 0.103$} & $0.829 \pm 0.012$ & $0.616 \pm 0.028$ \\
  & Token Probability &                                      & $0.704 \pm 0.165$ & $0.624 \pm 0.059$ \\
  & Sem.\ Uncertainty &                                      & $0.808 \pm 0.054$ & $0.642 \pm 0.018$ \\
\bottomrule
\end{tabular}
\end{table}

Table~\ref{tab:linguistic-auroc} then evaluates RALC's linguistic-space AUROC after rewriting, benchmarked against a Hedged QA baseline, a prompt-based black-box baseline to elicit hedged responses. Since RALC's quality is signal-dependent, stronger signals yield greater gains: semantic uncertainty consistently surpasses both the original AUROC and the Hedged QA baseline across all datasets, confirming that grounding rewritten expressions in a calibrated signal with principled retrieval-augmentation produces more discriminative outputs than black-box hedging.

\begin{table}[h!]
\centering
\caption{Linguistic-space AUROC before and after in-domain RALC with Hedged QA baseline, averaged across all five models. Original AUROC and Hedged QA AUROC are dataset-level quantities shared across all signals; post-RALC AUROC is signal-specific. Higher values indicate better discrimination between correct and incorrect responses in the linguistic space.}
\label{tab:linguistic-auroc}
\small
\begin{tabular}{llccc}
\toprule
\textbf{Dataset} & \textbf{Signal} & \textbf{Original AUROC} & \textbf{Post-RALC AUROC} & \textbf{Hedged QA AUROC} \\
\midrule
\multirow{3}{*}{MMLU}
  & Ling.\ Confidence & \multirow{3}{*}{$0.513 \pm 0.069$} & $0.533 \pm 0.019$ & \multirow{3}{*}{$0.565 \pm 0.020$} \\
  & Token Probability &                                      & $0.560 \pm 0.048$ &                                     \\
  & Sem.\ Uncertainty &                                      & $0.636 \pm 0.085$ &                                     \\
\cmidrule(lr){1-5}
\multirow{3}{*}{SQuAD~2.0}
  & Ling.\ Confidence & \multirow{3}{*}{$0.498 \pm 0.029$} & $0.488 \pm 0.011$ & \multirow{3}{*}{$0.521 \pm 0.044$} \\
  & Token Probability &                                      & $0.557 \pm 0.060$ &                                     \\
  & Sem.\ Uncertainty &                                      & $0.588 \pm 0.064$ &                                     \\
\cmidrule(lr){1-5}
\multirow{3}{*}{TruthfulQA}
  & Ling.\ Confidence & \multirow{3}{*}{$0.610 \pm 0.038$} & $0.600 \pm 0.034$ & \multirow{3}{*}{$0.627 \pm 0.028$} \\
  & Token Probability &                                      & $0.630 \pm 0.027$ &                                     \\
  & Sem.\ Uncertainty &                                      & $0.664 \pm 0.039$ &                                     \\
\bottomrule
\end{tabular}
\end{table}

\clearpage
\subsection{In-domain and cross-domain calibration results}
\label{appendix:cross-domain-calibration-value-chg}
Table~\ref{tab:cross-domain-calibration-value-chg} provides the absolute metric value changes corresponding to the percentage changes in Table~\ref{tab:cross-domain-calibration}, broken down by confidence estimator, training dataset, and evaluation dataset. Each entry reports the mean reduction in Faithfulness Divergence or generalised ECE across models (mean\,$\pm$\,std), with green indicating improvement and red indicating deterioration. In-domain results appear on the diagonal; off-diagonal entries reflect cross-domain transfer.

\begin{table}[h!]
\centering
\caption{In-domain and cross-domain linguistic-space calibration metric value changes for both Faithfulness Divergence and generalised ECE. We report the value change relative to the pre-calibration metrics (mean\,$\pm$\,std across models). Green text indicates calibration improvement (lower error), red indicates deterioration.}
\label{tab:cross-domain-calibration-value-chg}
\resizebox{\linewidth}{!}{
\begin{tabular}{p{2cm}p{2cm}lccc}
\toprule
\textbf{Metric} & \textbf{Signal} & \textbf{Train/Test} & {MMLU} & SQuAD~2.0 & {TruthfulQA} \\
\midrule
\multirow{9}{=}{\textbf{Faithfulness\\Divergence\\Mean\\Reduction}}
& \multirow{3}{=}{Linguistic\\Confidence}
& MMLU & \textcolor{green!70!black}{$\Delta$0.1057$\pm$0.1793} & \textcolor{green!70!black}{$\Delta$0.6384$\pm$0.2359} & \textcolor{green!70!black}{$\Delta$0.5436$\pm$0.2420} \\
& & SQuAD~2.0 & \textcolor{green!70!black}{$\Delta$0.3598$\pm$0.0779} & \textcolor{green!70!black}{$\Delta$1.1440$\pm$0.1398} & \textcolor{green!70!black}{$\Delta$1.0996$\pm$0.1566} \\
& & TruthfulQA & \textcolor{green!70!black}{$\Delta$0.2904$\pm$0.1946} & \textcolor{green!70!black}{$\Delta$1.2138$\pm$0.1430} & \textcolor{green!70!black}{$\Delta$1.1127$\pm$0.1695} \\
\cmidrule(lr){2-6}
& \multirow{3}{=}{Token\\Probability}
& MMLU & \textcolor{green!70!black}{$\Delta$0.0952$\pm$0.2094} & \textcolor{green!70!black}{$\Delta$0.9714$\pm$0.1862} & \textcolor{green!70!black}{$\Delta$1.0355$\pm$0.1556} \\
& & SQuAD~2.0 & \textcolor{green!70!black}{$\Delta$0.2768$\pm$0.0674} & \textcolor{green!70!black}{$\Delta$1.1987$\pm$0.1472} & \textcolor{green!70!black}{$\Delta$1.2931$\pm$0.1621} \\
& & TruthfulQA & \textcolor{green!70!black}{$\Delta$0.5249$\pm$0.0658} & \textcolor{green!70!black}{$\Delta$1.1848$\pm$0.1555} & \textcolor{green!70!black}{$\Delta$1.1890$\pm$0.1479} \\
\cmidrule(lr){2-6}
& \multirow{3}{=}{Semantic\\Uncertainty}
& MMLU & \textcolor{green!70!black}{$\Delta$0.2011$\pm$0.1424} & \textcolor{green!70!black}{$\Delta$0.6669$\pm$0.2484} & \textcolor{green!70!black}{$\Delta$0.7894$\pm$0.2090} \\
& & SQuAD~2.0 & \textcolor{green!70!black}{$\Delta$0.3700$\pm$0.0836} & \textcolor{green!70!black}{$\Delta$1.2180$\pm$0.1775} & \textcolor{green!70!black}{$\Delta$1.2042$\pm$0.1722} \\
& & TruthfulQA & \textcolor{green!70!black}{$\Delta$0.6437$\pm$0.0766} & \textcolor{green!70!black}{$\Delta$1.2246$\pm$0.1344} & \textcolor{green!70!black}{$\Delta$1.2160$\pm$0.1473} \\
\midrule
\multirow{9}{=}{\textbf{Generalised\\ECE\\Mean\\Reduction}}
& \multirow{3}{=}{Linguistic\\Confidence}
& MMLU & \textcolor{green!70!black}{$\Delta$0.0786$\pm$0.0242} & \textcolor{green!70!black}{$\Delta$0.0655$\pm$0.0258} & \textcolor{green!70!black}{$\Delta$0.0449$\pm$0.0244} \\
& & SQuAD~2.0 & \textcolor{green!70!black}{$\Delta$0.0394$\pm$0.0224} & \textcolor{green!70!black}{$\Delta$0.1372$\pm$0.0147} & \textcolor{green!70!black}{$\Delta$0.1286$\pm$0.0145} \\
& & TruthfulQA & \textcolor{green!70!black}{$\Delta$0.0725$\pm$0.0252} & \textcolor{green!70!black}{$\Delta$0.1492$\pm$0.0196} & \textcolor{green!70!black}{$\Delta$0.1462$\pm$0.0124} \\
\cmidrule(lr){2-6}
& \multirow{3}{=}{Token\\Probability}
& MMLU & \textcolor{green!70!black}{$\Delta$0.0966$\pm$0.0272} & \textcolor{green!70!black}{$\Delta$0.1217$\pm$0.0368} & \textcolor{green!70!black}{$\Delta$0.1381$\pm$0.0382} \\
& & SQuAD~2.0 & \textcolor{green!70!black}{$\Delta$0.0421$\pm$0.0063} & \textcolor{green!70!black}{$\Delta$0.1600$\pm$0.0158} & \textcolor{green!70!black}{$\Delta$0.2257$\pm$0.0245} \\
& & TruthfulQA & \textcolor{green!70!black}{$\Delta$0.1994$\pm$0.0167} & \textcolor{green!70!black}{$\Delta$0.1538$\pm$0.0194} & \textcolor{green!70!black}{$\Delta$0.1561$\pm$0.0147} \\
\cmidrule(lr){2-6}
& \multirow{3}{=}{Semantic\\Uncertainty}
& MMLU & \textcolor{green!70!black}{$\Delta$0.1028$\pm$0.0273} & \textcolor{green!70!black}{$\Delta$0.0974$\pm$0.0309} & \textcolor{green!70!black}{$\Delta$0.1240$\pm$0.0265} \\
& & SQuAD~2.0 & \textcolor{green!70!black}{$\Delta$0.0540$\pm$0.0323} & \textcolor{green!70!black}{$\Delta$0.1855$\pm$0.0295} & \textcolor{green!70!black}{$\Delta$0.1832$\pm$0.0258} \\
& & TruthfulQA & \textcolor{green!70!black}{$\Delta$0.2065$\pm$0.0184} & \textcolor{green!70!black}{$\Delta$0.1798$\pm$0.0251} & \textcolor{green!70!black}{$\Delta$0.1662$\pm$0.0165} \\
\bottomrule
\end{tabular}
}
\end{table}

\subsection{Further investigation on cross-domain calibration anomalies}
\label{sec:cross_domain_investigation}

Table~\ref{tab:cross-domain-calibration} reveals that cross-domain calibrators
occasionally outperform in-domain ones. We investigate this anomaly by examining the
miscalibration bias of each dataset, defined as the gap between mean expressed confidence
and mean accuracy, which determines how much signal is available for the calibration map
to learn from.

\begin{figure}[h!]
  \centering
  \includegraphics[width=\linewidth]{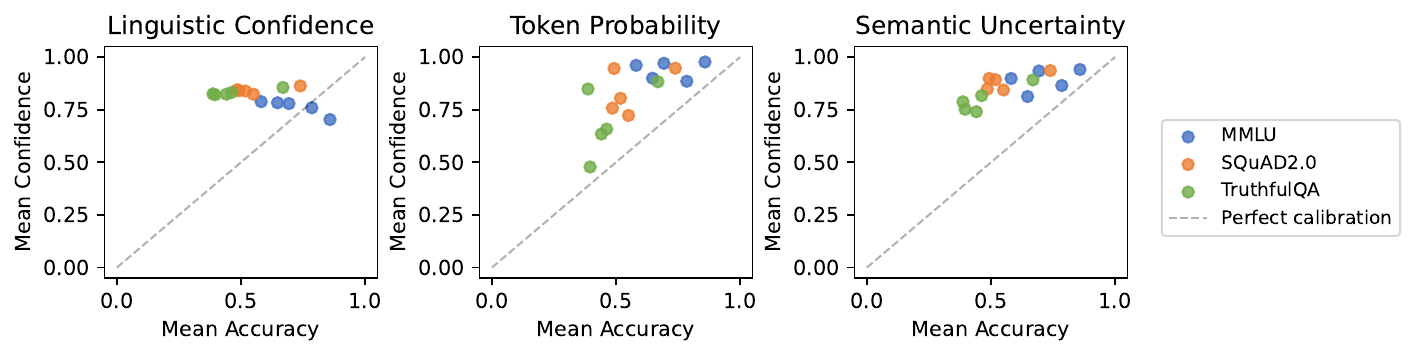}
  \caption{Mean confidence vs.\ mean accuracy per (dataset, model) pair. All datasets are
  systematically miscalibrated (above the diagonal), but the magnitude of bias varies
  considerably across domains.}
  \label{fig:bias}
\end{figure}

\begin{figure}[h!]
  \centering
  \includegraphics[width=\linewidth]{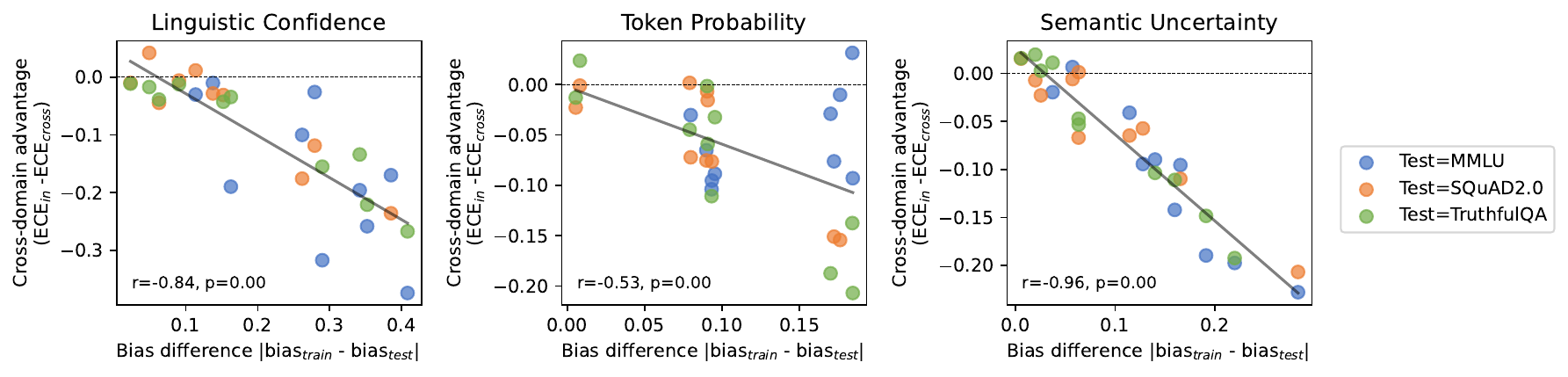}
  \caption{Miscalibration bias difference $|\bar{b}_\text{train} - \bar{b}_\text{test}|$
  vs.\ cross-domain advantage ($\text{ECE}_\text{in} - \text{ECE}_\text{cross}$). Colour indicates the test dataset. Transfer pairs with
  similar miscalibration biases achieve performance closer to in-domain calibration.}
  \label{fig:bias_corr}
\end{figure}

We measure the per-dataset miscalibration bias across confidence signals and models
(Table~\ref{tab:bias}; Figure~\ref{fig:bias}), and correlate the bias difference between
each source--target pair with the observed cross-domain advantage
(Figure~\ref{fig:bias_corr}).

\begin{table}[h!]
  \centering
  \small
  \caption{Mean miscalibration bias (mean confidence $-$ mean accuracy)
           per dataset and confidence signal, averaged over models.}
  \label{tab:bias}
  \begin{tabular}{lccc}
    \toprule
    Dataset        & Ling.\ Conf. & Token Prob. & Sem.\ Unc. \\
    \midrule
    MMLU           & $0.049$      & $0.225$     & $0.177$    \\
    SQuAD~2.0      & $0.285$      & $0.278$     & $0.326$    \\
    TruthfulQA     & $0.360$      & $0.229$     & $0.326$    \\
    \bottomrule
  \end{tabular}
\end{table}

All three datasets are systematically miscalibrated, but the magnitude differs
considerably. Figure~\ref{fig:bias_corr} shows a strong negative relationship between the source--target bias difference and the cross-domain advantage. Transfer pairs whose source and
target share a similar miscalibration bias show little performance gap relative to
in-domain calibration, whilst larger differences tend to favour in-domain calibration.

This pattern follows from the learning dynamics of the calibration map. When a target
domain has a weak bias, the in-domain calibrator has little signal to learn from and fits
an unreliable correction. A cross-domain source with a stronger, more consistent bias
learns a more decisive correction; provided the two domains share the same direction of
miscalibration, this correction transfers effectively even if its magnitude differs.
Cross-domain transfer therefore outperforms in-domain calibration precisely when
in-domain data is least informative.

\clearpage
\subsection{Detailed in-domain calibration vs. Hedged QA comparison}

Table~\ref{tab:detailed-in-domain-calibration-vs-hedged-qa-baseline} reports pre-to-post changes in linguistic-space Faithfulness Divergence and generalised ECE for RALC and the Hedged QA baseline, broken down by dataset, model, and confidence signal. Each cell shows the original Direct QA value alongside the post-intervention value; linguistic confidence is estimated by the LLM ensemble for all three response types. RALC outperforms Hedged QA in expectation across both metrics, with few exceptions due to model-specific signal characteristics, whilst Hedged QA shows limited and inconsistent improvements.

\begin{table*}[ht]
\centering
\caption{Detailed in-domain calibration vs. Hedged QA comparison.}
\label{tab:detailed-in-domain-calibration-vs-hedged-qa-baseline}
\resizebox{\textwidth}{!}{%
\begin{tabular}{llllcccc}
\toprule
\multirow{3}{*}{Dataset}
  & \multirow{3}{*}{Model}
  & \multirow{3}{*}{Signal}
  & \multicolumn{2}{c}{\textbf{In-Domain Calibration}}
  & \multicolumn{2}{c}{\textbf{Hedged QA}} \\
\cmidrule(lr){4-5} \cmidrule(lr){6-7}
  & & & \makecell{Faithfulness \\ Divergence \\ (Orig.\ $\to$ Calib.)}
       & \makecell{Generalised \\ ECE \\ (Orig.\ $\to$ Calib.)}
       & \makecell{Faithfulness \\ Divergence \\ (Orig.\ $\to$ Hedged)}
       & \makecell{Generalised \\ ECE \\ (Orig.\ $\to$ Hedged)} \\
\midrule
 \multirow{15}{*}{MMLU} & \multirow{3}{*}{Mistral-7B-Inst.} & Linguistic Conf. & $1.198 \to 0.557$ & $0.239 \to 0.120$ & \multirow{3}{*}{$1.198 \to 1.227$} & \multirow{3}{*}{$0.239 \to 0.251$} \\
 & & Token Prob. & $1.198 \to 0.578$ & $0.239 \to 0.109$ &  &  \\
 & & Semantic Unc. & $1.198 \to 0.471$ & $0.239 \to 0.113$ &  &  \\
\cmidrule(l){2-7}
 & \multirow{3}{*}{Gemma-4-31B-IT} & Linguistic Conf. & $1.047 \to 1.077$ & $0.284 \to 0.124$ & \multirow{3}{*}{$1.047 \to 0.577$} & \multirow{3}{*}{$0.284 \to 0.146$} \\
 & & Token Prob. & $1.047 \to 0.956$ & $0.284 \to 0.093$ &  &  \\
 & & Semantic Unc. & $1.047 \to 0.530$ & $0.284 \to 0.080$ &  &  \\
\cmidrule(l){2-7}
 & \multirow{3}{*}{Llama-3.1-8B-Inst.} & Linguistic Conf. & $0.925 \to 0.602$ & $0.190 \to 0.120$ & \multirow{3}{*}{$0.925 \to 0.961$} & \multirow{3}{*}{$0.190 \to 0.199$} \\
 & & Token Prob. & $0.925 \to 0.665$ & $0.190 \to 0.096$ &  &  \\
 & & Semantic Unc. & $0.925 \to 0.461$ & $0.190 \to 0.092$ &  &  \\
\cmidrule(l){2-7}
 & \multirow{3}{*}{GPT-OSS-20B} & Linguistic Conf. & $0.579 \to 1.118$ & $0.152 \to 0.128$ & \multirow{3}{*}{$0.579 \to 0.530$} & \multirow{3}{*}{$0.152 \to 0.146$} \\
 & & Token Prob. & $0.579 \to 1.333$ & $0.152 \to 0.138$ &  &  \\
 & & Semantic Unc. & $0.579 \to 0.563$ & $0.152 \to 0.089$ &  &  \\
\cmidrule(l){2-7}
 & \multirow{3}{*}{Qwen3-8B} & Linguistic Conf. & $0.902 \to 0.770$ & $0.167 \to 0.148$ & \multirow{3}{*}{$0.902 \to 0.851$} & \multirow{3}{*}{$0.167 \to 0.160$} \\
 & & Token Prob. & $0.902 \to 0.643$ & $0.167 \to 0.113$ &  &  \\
 & & Semantic Unc. & $0.902 \to 0.564$ & $0.167 \to 0.145$ &  &  \\
\midrule
 \multirow{15}{*}{TruthfulQA} & \multirow{3}{*}{Mistral-7B-Inst.} & Linguistic Conf. & $1.782 \to 0.710$ & $0.374 \to 0.227$ & \multirow{3}{*}{$1.782 \to 1.654$} & \multirow{3}{*}{$0.374 \to 0.350$} \\
 & & Token Prob. & $1.782 \to 0.727$ & $0.374 \to 0.248$ &  &  \\
 & & Semantic Unc. & $1.782 \to 0.477$ & $0.374 \to 0.238$ &  &  \\
\cmidrule(l){2-7}
 & \multirow{3}{*}{Gemma-4-31B-IT} & Linguistic Conf. & $1.272 \to 0.831$ & $0.209 \to 0.118$ & \multirow{3}{*}{$1.272 \to 1.271$} & \multirow{3}{*}{$0.209 \to 0.204$} \\
 & & Token Prob. & $1.272 \to 0.625$ & $0.209 \to 0.100$ &  &  \\
 & & Semantic Unc. & $1.272 \to 0.469$ & $0.209 \to 0.096$ &  &  \\
\cmidrule(l){2-7}
 & \multirow{3}{*}{Llama-3.1-8B-Inst.} & Linguistic Conf. & $1.828 \to 0.712$ & $0.427 \to 0.269$ & \multirow{3}{*}{$1.828 \to 1.756$} & \multirow{3}{*}{$0.427 \to 0.394$} \\
 & & Token Prob. & $1.828 \to 0.636$ & $0.427 \to 0.247$ &  &  \\
 & & Semantic Unc. & $1.828 \to 0.467$ & $0.427 \to 0.254$ &  &  \\
\cmidrule(l){2-7}
 & \multirow{3}{*}{GPT-OSS-20B} & Linguistic Conf. & $2.265 \to 0.678$ & $0.392 \to 0.230$ & \multirow{3}{*}{$2.265 \to 2.143$} & \multirow{3}{*}{$0.392 \to 0.375$} \\
 & & Token Prob. & $2.265 \to 0.645$ & $0.392 \to 0.227$ &  &  \\
 & & Semantic Unc. & $2.265 \to 0.475$ & $0.392 \to 0.179$ &  &  \\
\cmidrule(l){2-7}
 & \multirow{3}{*}{Qwen3-8B} & Linguistic Conf. & $2.054 \to 0.707$ & $0.442 \to 0.270$ & \multirow{3}{*}{$2.054 \to 1.999$} & \multirow{3}{*}{$0.442 \to 0.413$} \\
 & & Token Prob. & $2.054 \to 0.623$ & $0.442 \to 0.242$ &  &  \\
 & & Semantic Unc. & $2.054 \to 0.464$ & $0.442 \to 0.247$ &  &  \\
\midrule
 \multirow{15}{*}{SQuAD~2.0} & \multirow{3}{*}{Mistral-7B-Inst.} & Linguistic Conf. & $1.978 \to 0.701$ & $0.350 \to 0.185$ & \multirow{3}{*}{$1.978 \to 1.943$} & \multirow{3}{*}{$0.350 \to 0.343$} \\
 & & Token Prob. & $1.978 \to 0.682$ & $0.350 \to 0.181$ &  &  \\
 & & Semantic Unc. & $1.978 \to 0.487$ & $0.350 \to 0.116$ &  &  \\
\cmidrule(l){2-7}
 & \multirow{3}{*}{Gemma-4-31B-IT} & Linguistic Conf. & $1.288 \to 0.649$ & $0.166 \to 0.094$ & \multirow{3}{*}{$1.288 \to 1.300$} & \multirow{3}{*}{$0.166 \to 0.167$} \\
 & & Token Prob. & $1.288 \to 0.593$ & $0.166 \to 0.074$ &  &  \\
 & & Semantic Unc. & $1.288 \to 0.480$ & $0.166 \to 0.103$ &  &  \\
\cmidrule(l){2-7}
 & \multirow{3}{*}{Llama-3.1-8B-Inst.} & Linguistic Conf. & $1.594 \to 0.643$ & $0.305 \to 0.160$ & \multirow{3}{*}{$1.594 \to 1.584$} & \multirow{3}{*}{$0.305 \to 0.302$} \\
 & & Token Prob. & $1.594 \to 0.623$ & $0.305 \to 0.145$ &  &  \\
 & & Semantic Unc. & $1.594 \to 0.495$ & $0.305 \to 0.122$ &  &  \\
\cmidrule(l){2-7}
 & \multirow{3}{*}{GPT-OSS-20B} & Linguistic Conf. & $2.265 \to 0.764$ & $0.389 \to 0.232$ & \multirow{3}{*}{$2.265 \to 2.214$} & \multirow{3}{*}{$0.389 \to 0.381$} \\
 & & Token Prob. & $2.265 \to 0.665$ & $0.389 \to 0.192$ &  &  \\
 & & Semantic Unc. & $2.265 \to 0.474$ & $0.389 \to 0.137$ &  &  \\
\cmidrule(l){2-7}
 & \multirow{3}{*}{Qwen3-8B} & Linguistic Conf. & $2.103 \to 0.751$ & $0.370 \to 0.225$ & \multirow{3}{*}{$2.103 \to 2.033$} & \multirow{3}{*}{$0.370 \to 0.360$} \\
 & & Token Prob. & $2.103 \to 0.670$ & $0.370 \to 0.189$ &  &  \\
 & & Semantic Unc. & $2.103 \to 0.480$ & $0.370 \to 0.176$ &  &  \\
\bottomrule
\end{tabular}%
}
\end{table*}

\clearpage
\subsection{In-domain calibration reliability diagrams}
\begin{figure}[h!]
    \centering
    \includegraphics[width=0.8\linewidth]{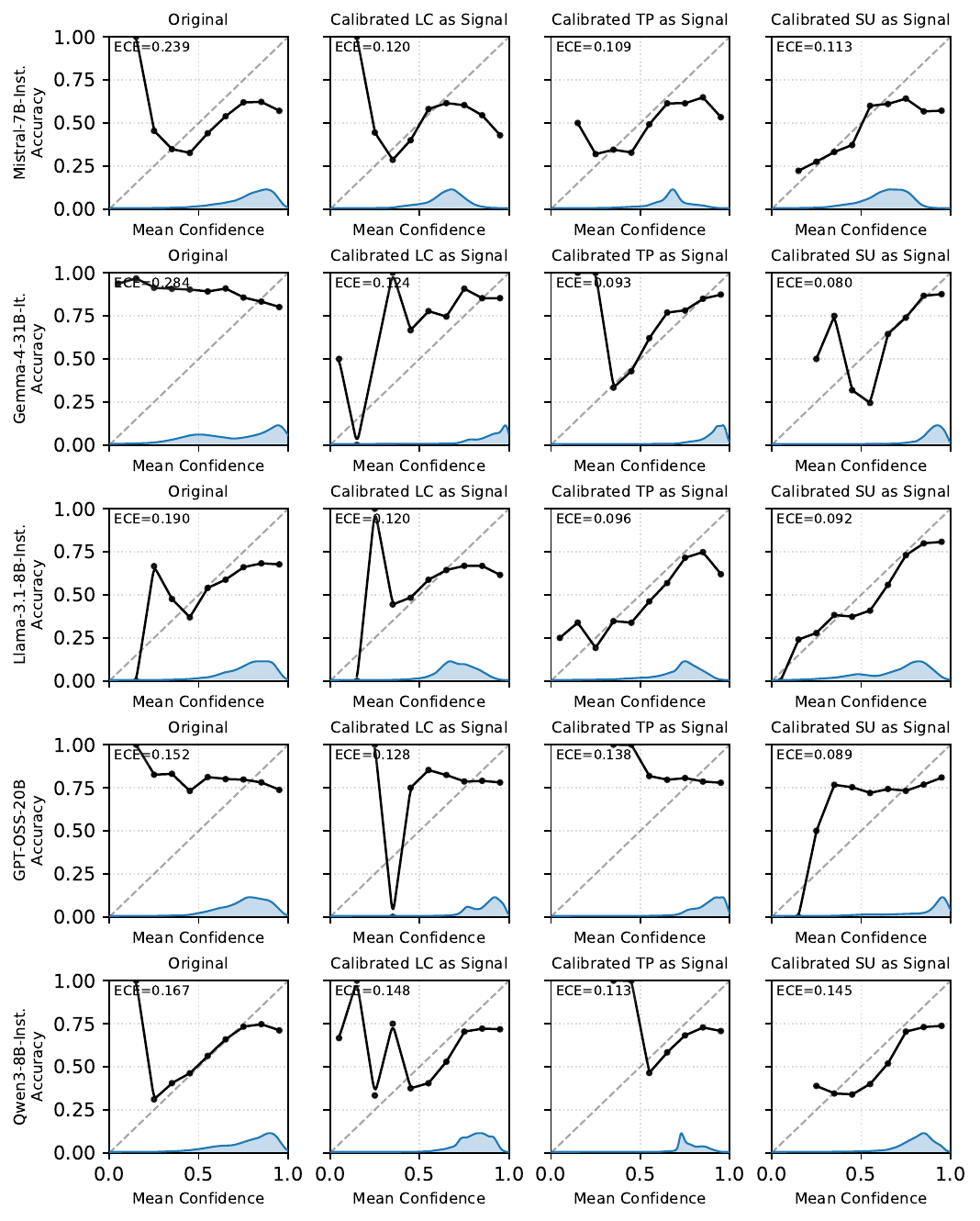}
    \caption{In-domain calibration reliability diagrams for MMLU across confidence signals and models. The left column shows the original linguistic confidence of the Direct QA responses. The rest of the columns show the linguistic confidence of the rewritten responses through RALC guided by different confidence signals, including linguistic confidence (LC), token probability (TP), and semantic uncertainty (SU). We report the generalised ECE \citep{wang2025calibratingexpressionscertainty} along with the reliability diagrams. The results show that RALC effectively reduces miscalibration across all confidence signals and models.}
    \label{fig:in-domain-reliability-diagrams-mmlu}
\end{figure}

\begin{figure}[h!]
    \centering
    \includegraphics[width=0.8\linewidth]{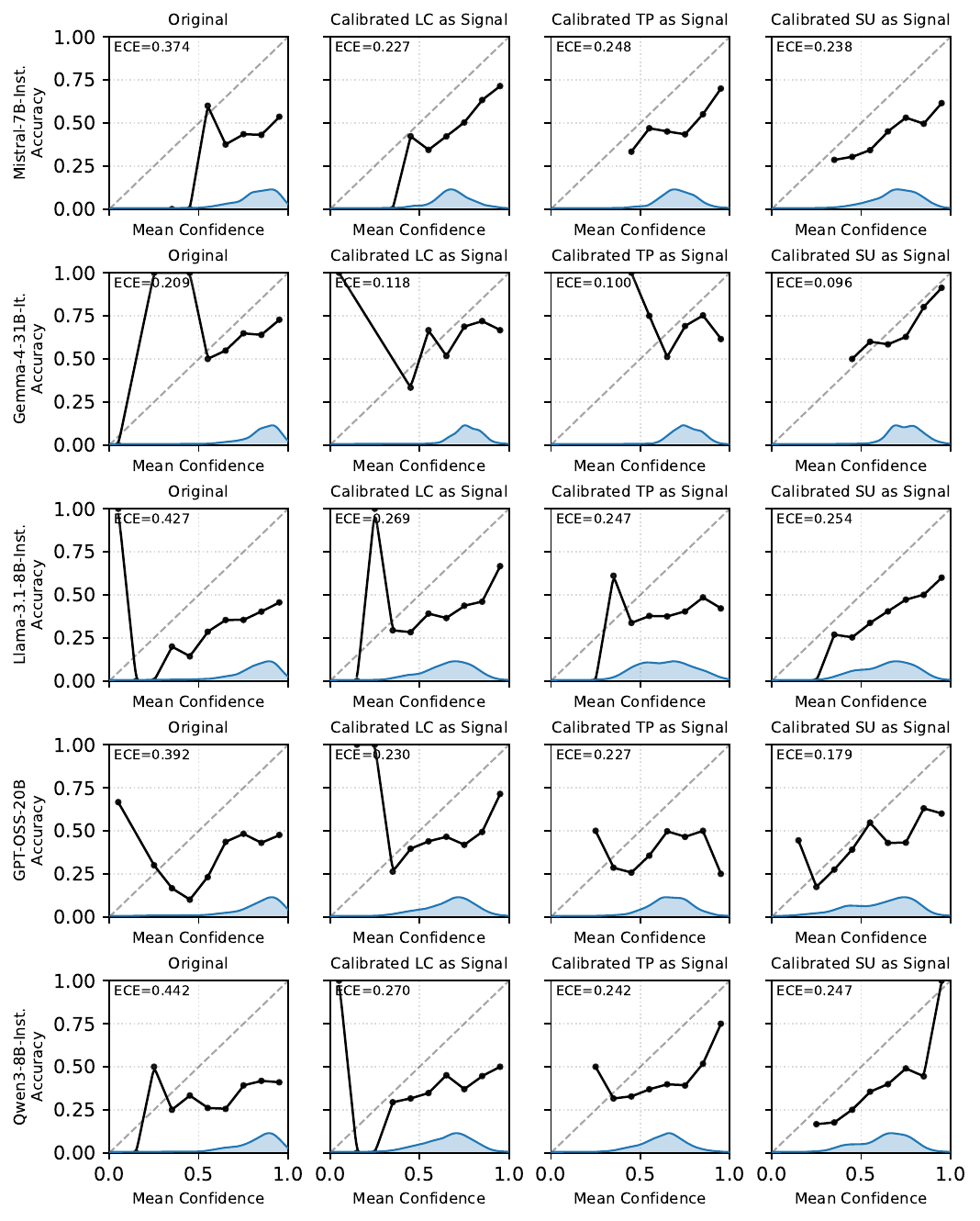}
    \caption{In-domain calibration reliability diagrams for TruthfulQA across confidence signals and models. The left column shows the original linguistic confidence of the Direct QA responses. The rest of the columns show the linguistic confidence of the rewritten responses through RALC guided by different confidence signals, including linguistic confidence (LC), token probability (TP), and semantic uncertainty (SU). We report the generalised ECE \citep{wang2025calibratingexpressionscertainty} along with the reliability diagrams. The results show that RALC effectively reduces miscalibration across all confidence signals and models.}
    \label{fig:in-domain-reliability-diagrams-truthfulqa}
\end{figure}

Figures~\ref{fig:in-domain-reliability-diagrams-mmlu} and \ref{fig:in-domain-reliability-diagrams-truthfulqa} present the in-domain calibration reliability diagrams for MMLU and TruthfulQA, respectively, across confidence signals and models. The left column shows the original linguistic confidence of the Direct QA responses. The rest of the columns show the linguistic confidence of the rewritten responses through RALC guided by different confidence signals, including linguistic confidence (LC), token probability (TP), and semantic uncertainty (SU). Additionally, we report the generalised ECE \citep{wang2025calibratingexpressionscertainty} and Faithfulness Divergence (FD) along with the reliability diagrams. The results show that RALC effectively reduces miscalibration across all confidence signals and models. 

\section{LLM configurations}
\label{appendix:llm-config}
All five evaluation targets, the LLM ensemble, and the LLM rewriter in the RALC pipeline are configured with a temperature of 1 to encourage diverse outputs. The LLM cluster selector and grader are configured with a temperature of 0 to encourage deterministic outputs. All models are hosted locally (single RTX 4090 GPU) or through cloud APIs.


\end{document}